\documentclass[10pt,twocolumn,letterpaper]{article}

\usepackage[pagenumbers]{cvpr}      % To produce the REVIEW version
\usepackage{graphicx}
\usepackage{amsmath}
\usepackage{amssymb}
\usepackage{booktabs}
\usepackage{soul}

\usepackage{makecell}
\usepackage{tablefootnote}
\usepackage{multirow}

\usepackage[pagebackref,breaklinks,colorlinks]{hyperref}

\usepackage[capitalize]{cleveref}
\crefname{section}{Sec.}{Secs.}
\Crefname{section}{Section}{Sections}
\Crefname{table}{Table}{Tables}
\crefname{table}{Tab.}{Tabs.}

\def\cvprPaperID{8} % *** Enter the Biometrics Workshop CVPR Paper ID here
\def\confName{CVPR 2023 Biometrics Workshop}
\def\confYear{2023}

\begin{document}

\title{PIC-Score: Probabilistic Interpretable Comparison Score for Optimal Matching Confidence in Single- and
Multi-Biometric Face Recognition}

\author{Pedro C. Neto\textsuperscript{1,2}, Ana F. Sequeira\textsuperscript{1,2}, Jaime S. Cardoso\textsuperscript{1,2} and Philipp Terh\"{o}rst\textsuperscript{3}\\
\textsuperscript{1}INESC TEC, Porto, Portugal\\
\textsuperscript{2}Faculdade de Engenharia da Universidade do Porto, Porto, Portugal\\
\textsuperscript{3}Paderborn University, Paderborn, Germany}
\maketitle
%Paderborn University\\
%%%%%%%%% ABSTRACT
\begin{abstract}

In the context of biometrics, matching confidence refers to the confidence that a given matching decision is correct.
Since many biometric systems operate in critical decision-making processes, such as in forensics investigations, accurately and reliably stating the matching confidence becomes of high importance.
Previous works on biometric confidence estimation can well differentiate between high and low confidence, but lack interpretability.
Therefore, they do not provide accurate probabilistic estimates of the correctness of a decision.
In this work, we propose a probabilistic interpretable comparison (PIC) score that accurately reflects the probability that the score originates from samples of the same identity.
We prove that the proposed approach provides optimal matching confidence. 
Contrary to other approaches, it can also optimally combine multiple samples in a joint PIC score which further increases the recognition and confidence estimation performance.
In the experiments, the proposed PIC approach is compared against all biometric confidence estimation methods available on four publicly available databases and five state-of-the-art face recognition systems.
The results demonstrate that PIC has a significantly more accurate probabilistic interpretation than similar approaches and is highly effective for multi-biometric recognition.
The code is publicly-available\footnote{\url{https://github.com/pterhoer/OptimalMatchingConfidence} 
%\scriptsize\url{https://github.com/pterhoer/OptimalMatchingConfidence}
}.

\end{abstract}

%%%%%%%%% Introduction
\section{Introduction}
\label{sec:Introduction}

Biometric recognition systems, such as face recognition, have a growing effect on our daily life \cite{DBLP:journals/ijon/WangD21a}.
Since these systems are increasingly involved in critical decision-making processes, such as in forensics and law enforcement, it is important for these applications to act on reliable decisions \cite{DBLP:journals/corr/abs-2210-10354,neto2022explainable}.
While human operators can intuitively state how sure they are about a decision and if they can carry out justifiable actions based on this decision \cite{yeung2012metacognition}, current biometric systems do not possess such reliable confidence estimates.

\begin{figure}[th]
    \centering
    \includegraphics[width=0.47\textwidth]{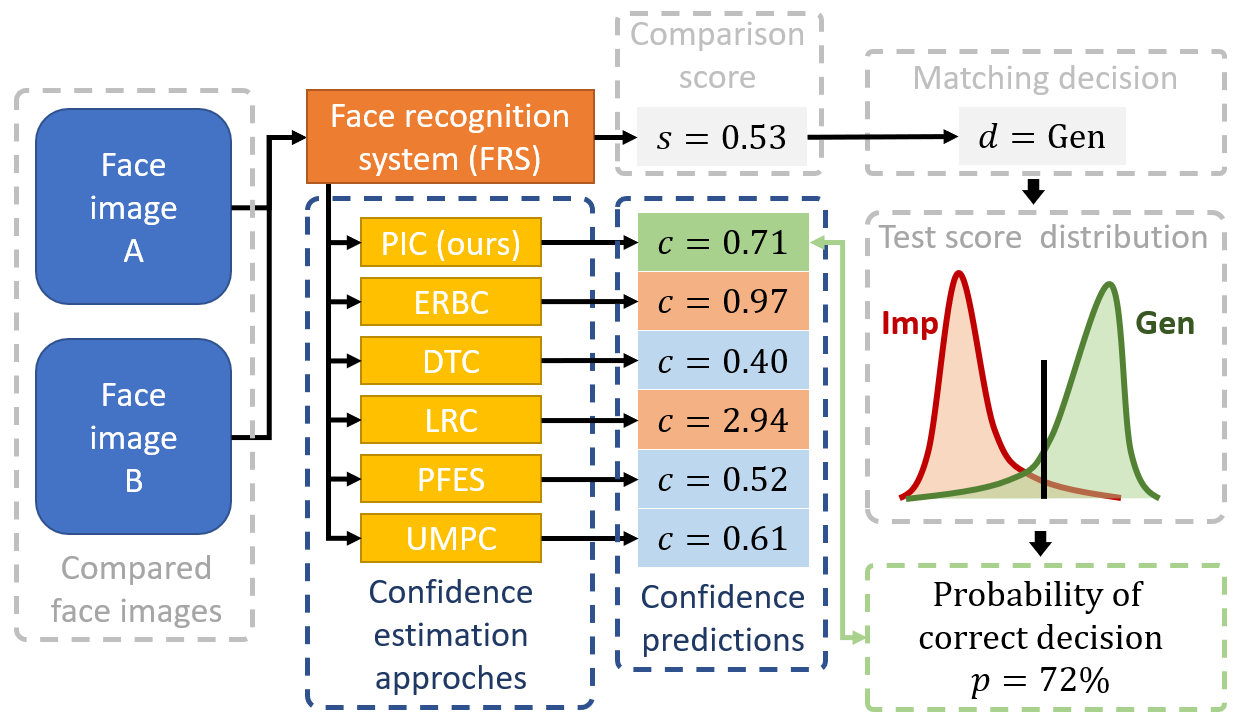}
    \caption{\textbf{Probabilistic Confidence Interpretation Problem} - Given two biometric samples, the face recognition system comes to the decision that the sample belonging to the same identity (genuine). Evaluating the correctness of this decision based on independent test data results in a probability of 72\% that this decision is correct. However, most confidence estimation approaches (yellow) either overestimate (red) or underestimate (blue) the confidence. Contrarily, the proposed PIC approach provides a confidence value (green) reflecting the probability the decision is correct. 
    }
    \label{fig:Visualization}
\end{figure}

For face recognition (FR), to decide if two faces belong to the same person, a feature representation for each sample is created by a face recognition model.
These are known as templates and comparing two templates with a similarity function, such as cosine similarity, results in a comparison score that describes the similarity between the two faces.
If the comparison score is above a given threshold, the matching decision is genuine (same identity).
Otherwise, the decision is imposter (different identities) \cite{10.5555/1952072}.

The score uncertainty describes the uncertainty in the score depending on the uncertainty of the data and the model.
Contrarily, decision confidence refers to the confidence that the made decision is correct \cite{DBLP:journals/corr/abs-2210-10354, peterson1988confidence}.
A low-confidence decision is therefore more likely to be wrong than a high-confident one.
Consequently, confidence estimation might prevent high-cost mistakes e.g. in front of a court.
Previous works proposed several approaches to state the confidence of a model's decision.
However, these confidence measures lack interpretability (see Figure \ref{fig:Visualization}), meaning that those are hardly interpretable and thus, do not reflect the probability that the matching decision is correct.

In this work, we propose a probabilistic interpretable comparison (PIC) score that accurately reflects the probability that the score originates from samples of the same identity.
Additionally, the PIC score provides a natural way of combing several comparisons from multiple samples originated from the same network without losing its probabilistic interpretability.
The experiments were conducted on five state-of-the-art face recognition systems (FRS) and four publicly-available datasets.
Comparing the proposed approach against all available biometric confidence estimation methods, the results demonstrate that PIC results in much more accurate, stable, and interpretable confidence estimates.
Moreover, PIC scores from multiple instances lead to a strong boost in recognition performance and probabilistic interpretability.

In contrast to previous works, the proposed PIC score unifies several beneficial properties:
\begin{itemize}
    \item \textbf{Interpretability:} The PIC score accurately reflects the probability that the comparison belongs to a genuine (same person) comparison. This allows critical decision-making processes to act upon reliable decisions or, in case of low confidence, allow to ask a more confident system or human operator for the decision.
    \item \textbf{Optimality:} Despite its simplicity, PIC is derived from Bayes' theorem and thus, provides optimal matching confidences given suitable training data. This might be useful, e.g. in law enforcement, when approaches with a theoretical foundation are preferred.
    \item \textbf{Universality:} Since PIC operates on score level, it is not limited to face and can be applied to any biometric modality and recognition model without changing its single-biometric performance. 
    \item \textbf{Combinability:} Contrarily to standard comparison scores, a joint PIC score can naturally be computed when multiple samples are given. This leads to stronger multi-biometric recognition performance without losing its interpretability and is highly beneficial when e.g. dealing with multiple video frames.
    \item \textbf{Integratability:} PIC is easily integrateable. It can be easily added on top of an existing biometric system without retraining that system. Moreover, it avoids the need for data- and computationally-expensive experiments to determine the wanted decision threshold due to its interpretable nature.
\end{itemize}

%%%%%%%%% Related work
\section{Related Work}
\label{sec:RelatedWork}

Confidence estimation in biometric recognition is a relatively new but important field.
It aims at estimating the confidence that a decision is correct \cite{DBLP:journals/corr/abs-2210-10354, peterson1988confidence}.
{While for biometric attribute estimation} \cite{DBLP:conf/btas/TerhorstHKZDKK19} model calibration methods for classification \cite{DBLP:conf/icml/GuoPSW17} can be easily adapted due to the output of probability values, this becomes more challenging for zero-shot representation learning tasks, such as biometric recognition.
%Similarly, confidence estimation methods so far focus on the modality face.
To the best of our knowledge, there are only five recent confidence estimation methods for face recognition.
In \cite{DBLP:conf/btas/ZeinstraMVS18}, Zeinstra et al. proposed to use the likelihood ratio between genuine and imposter as a confidence measure for forensic use cases.
Huber et al. \cite{DBLP:journals/corr/abs-2210-10354} proposed a more intuitive, but effective solution, by utilizing the distance between the score and the decision threshold as a confidence measure.
To get probabilistic confidence estimates, in \cite{huberICPR22}, the corresponding error rate for the decision threshold, lying closest to a given comparison score, was interpreted as the decision confidence.
Other approaches \cite{DBLP:journals/corr/abs-2210-10354,DBLP:conf/iccv/ShiJ19} require a specialized network that produces probabilistic face embeddings with corresponding feature uncertainties.
In \cite{DBLP:journals/corr/abs-2210-10354}, Huber et al. propagated these uncertainties through an approximated decision function to obtain matching confidence.
In \cite{DBLP:conf/iccv/ShiJ19}, Shi and Jain used these uncertainties to compute the probability that two samples share the same face embedding.

Table \ref{tab:PropertiesSOTA} compares the properties of existing biometric confidence estimation methods.
Only the proposed PIC approach is jointly (probabilistic) interpretable, optimal, universal, combinable, and integrable.
While all of these confidence estimation methods work well in differentiating between lower and higher confidence, our experiments will demonstrate that the produced confidence estimates can not be well interpreted as the probability that a decision is correct.
To fill this gap, we propose the PIC score approach.

\begin{table}[ht]
\renewcommand{\arraystretch}{1.0}
\setlength{\tabcolsep}{2.8pt}
    \centering
    \footnotesize
    \caption{Properties of biometric confidence estimation approaches.}
    \label{tab:PropertiesSOTA}
    \begin{tabular}{lccccc}
    \Xhline{2\arrayrulewidth}
         Method & Interpretable & Optimal & Universal & Combinable & Integrable \\
         \midrule
         DTC \cite{DBLP:journals/corr/abs-2210-10354} & $-$& $-$& $+$& $-$& $+$\\
         LRC \cite{DBLP:conf/btas/ZeinstraMVS18} & $-$& $-$& $+$& $+$& $+$\\
         ERBC \cite{huberICPR22} & $+$& $-$& $+$& $+$& $+$\\
         PFES \cite{DBLP:conf/iccv/ShiJ19} & $+$& $-$& $-$& $+$& $-$\\
         UPMC \cite{DBLP:journals/corr/abs-2210-10354} & $-$& $-$& $-$& $-$& $-$\\
         PIC (Ours) & $+$ & $+$ & $+$ & $+$ & $+$ \\
    \Xhline{2\arrayrulewidth}
    \end{tabular}
\end{table}

%%%%%%%%% Methodology
\section{Methodology}
\label{sec:Methodology}

\subsection{Probabilistic Interpretable Comparison Score}
\label{sec:Methodolgy-PIC}
To make the PIC score probabilistic interpretable, we define the score $s_{PIC}(\bar{s}) = P(g(s)|\bar{s})$ as the probability that the set of comparison scores $\bar{s}$ originates from the genuine distribution $g(s)$.
To be precise, given are $n$ distributed standard comparison scores $\bar{s}=\{s_1, s_2, \dots s_n\}$ and we want to compute the probability $P(g(s)|\bar{s})$ that these scores were drawn from the genuine distribution $g(s)$ rather from the imposter distributions $f(s)$.
The probability can be modeled by the Bayes rule
\begin{align}
    P(g(s)|\bar{s}) = \dfrac{P(\bar{s}|g(s)) \cdot P(g(s))}{P(\bar{s}|g(s)) \cdot P(g(s)) + P(\bar{s}|f(s)) \cdot P(f(s))}
\end{align} 
where $P(g(s))$ and $P(f(s))$ are the prior probabilities that the scores belong to the genuine or imposter distribution.
Assuming that the scores are drawn independently, the likelihood functions $P(\bar{s}|g(s))$ and $P(\bar{s}|f(s))$ are given by
\begin{align}
    P(\bar{s}|g(s)) &= L_g(\bar{s}) = g(s_1) \cdot g(s_2)  \dots g(s_n) \\
    P(\bar{s}|f(s)) &= L_f(\bar{s}) = f(s_1) \cdot f(s_2)  \dots f(s_n).
\end{align}
For simplicity, we further assume that genuine and imposter comparisons happen equally often $P(g(s))=P(f(s))$ and obtain the multi-instance biometric solution $P(g(s)|\bar{s})$ that we define as the PIC score
\begin{align}
    s_{PIC}(\bar{s}) = P(g(s)|\bar{s}) = \dfrac{L_g(\bar{s})}{L_g(\bar{s}) + L_f(\bar{s})}. \label{eq:SimplifiedEquation}
\end{align}
For a single-instance scenario ($\bar{s} = s_1$), this simplifies to
\begin{align}
     s_{PIC}(s_1) = P(g(s)|s_1) = \dfrac{g(s_1)}{g(s_1) + f(s_1)}.
\end{align}
If a system decides for genuine, $s_{PIC}(s_1) = P(g(s)|s_1)$ is the probability that the decision is correct.
Contrarily, if the decision is imposter, then the probability for a correct decision is given by $1-s_{PIC}(s_1)$. 
{The simplification assumption of equal prior probabilities aims at keeping the confidence unbiased by having an equalized weighting between both kinds of decision errors.
However, it can be optimized for particular applications (e.g. border control) by specifying these prior probabilities to operational conditions.}

\subsection{Training PIC}
So far, the derivation assumes that we know $g(s)$ and $f(s)$ in advance.
The training processes of PIC involve learning these probability density distributions, e.g. with kernel density estimation (KDE). 
Given the training data $\mathcal{D}\in \{s_i, y_i\}_{i=1,\dots,N}$ consisting of pairs with comparison scores $\{s_i\}$ and if these belong to genuine or imposter comparisons, the training data scores are split in genuine and imposter scores $\mathcal{D}_{g}$ and $\mathcal{D}_{f}$.
Then, for each of the score sets a probability density distribution is learned via KDE. For genuine, this is given by
\begin{align}
    g(s) = \dfrac{1}{|\mathcal{D}_{g}| \cdot h} \sum_{s_i \in \mathcal{D}_g} K \left( \dfrac{s-s_i}{h} \right). 
\end{align}
As the kernel $K(x)$, we used a Gaussian Kernel
\begin{align}
    K(x) = \frac{1}{\sqrt{2 \pi}} e ^{\frac{-x^2}{2}}
\end{align}
and selected the bandwidth $h$ with the Scott's rule for one dimension $h=N^{-\frac{1}{5}}$.
These resulted in training the probability density distributions $f(s)$ and $g(s)$ needed for the PIC score calculation.

\subsection{Discussion}
A comparison score describes the similarity between two samples.
A higher score refers to a higher chance of belonging to the same identity and vice versa.
From this perspective, it makes sense to interpret the probability of samples originating from the genuine distribution (or not) as a comparison score.
For a single comparison, as long as $\frac{f(s)}{g(s)}$ is monotonic, the order of PICS scores, and thus its single-comparison recognition performance is identical to the ones of the standard comparison score.
However, since the derivation of $s_{PIC}(\bar{s}) = P(g(s)|\bar{s})$ was already done for the case of multiple comparisons, its probabilistic interpretation, and the fusion process is optimal.
Since inference with Gaussian KDE can be slow with large training data, and thus the PIC score calculation, we recommend creating look-up tables for $g(s)$ and $f(s)$ to speed up the score calculation.
Lastly, the PIC scores avoid the need for data- and computationally-expensive experiments to determine the wanted decision threshold.
Since the scores already reflect the probabilities for errors, the threshold $t$ for a false match rate (FMR) can be chosen by $t=1-\text{FMR}$.

%%%%%%%%% Experimental setup
\section{Experimental Setup} 
\label{sec:ExperimentalSetup}

\subsection{Databases}
The experiments were conducted on four publicly-available face recognition datasets with various properties.
The Adience dataset \cite{Eidinger:2014:AGE:2771306.2772049} consists of 26k images from over 2k different subjects.
The images of the Adience dataset possess a wide range of challenges such as low image quality and very young faces.
LFW is a dataset \cite{LFWTech} containing 13k face images of over 5k identities that were captioned from news images.
The ColorFeret database \cite{DBLP:journals/pami/PhillipsMRR00} consists of 14k face images from over 1k different individuals with a variety of face poses (from frontal to profile) and facial expressions under well-controlled conditions.
Lastly, the Morph \cite{DBLP:conf/fgr/RicanekT06} dataset consists of 55k frontal face images from over 13k subjects in high resolution.
For training, we applied a subject-exclusive test (50\%) train split (50\%). The 50/50\% train split is respective to the number of genuine and impostor samples. As such, the identities of both sets are selected in a way that the sum of the combination of samples per identity is similar in both.
Since we don't want to add prior knowledge about the evaluation process in the training step, we use the simplified version of the PIC score with $P(g(s))=P(f(s))$ as described in Equation \ref{eq:SimplifiedEquation}.

\subsection{Evaluation Metrics}
For evaluating the recognition performance, we follow the international standard for biometric verification evaluation \cite{ISO_Metrik} by reporting the face verification error in terms of false non-match rate (FNMR) at fixed false match rate (FMR).
In the experiments, we focus on reporting the FNMR at $10^{-3}$ FMR as recommended by the best practice guidelines for automated border control of the European Border Guard Agency Frontex \cite{FrontexBestPractice}.

To evaluate the probabilistic interpretability of the confidence scores, we adapt the widely-used Expected Calibration Error (ECE) and Maximum Calibration Error (MCE) \cite{DBLP:conf/aaai/NaeiniCH15, DBLP:conf/icml/GuoPSW17}.
{In contrast to error-vs-reject curves from quality assessment} \cite{DBLP:conf/cvpr/TerhorstKDKK20} which evaluates the order of confidence prediction and neglects their probabilistic interpretation, this work focuses on the widely-used confidence estimation metrics ECE and MCE.
Dividing $n$ samples into $M$ equally-spaced bins $B_m$ based on their confidence, the ECE
\begin{align}
ECE = \sum_{m=1}^M \dfrac{|B_m|}{n} \,\, \lvert p_{true}(B_m) - p_{pred}(B_m)  \rvert
\end{align}
describes the average error between the model's predicted confidence $p_{pred}$ and its true confidence $p_{true}$ given the test data.
Since reliable confidence estimates are absolutely necessary in high-security applications, the MCE
\begin{align}
MCE = \max_{m\in\{1,...,M\}} \lvert p_{true}(B_m) - p_{pred}(B_m)  \rvert
\end{align}
measures the worst-case deviation between the predicted and true confidence.
For a perfect confidence estimator and suitable test data, the MCE and ECE are both zero.

\subsection{Face Recognition Models}
To ensure high compatibility of confidence estimation methods with a wide variety of recognition systems, the experiments were conducted on five state-of-the-art face recognition models pre-trained and released by the corresponding authors\footnote{For FaceNet, the authors never made the model publicly-available. Instead, a third-party implementation was used: \url{https://github.com/davidsandberg/facenet}}.
This includes the models FaceNet \cite{DBLP:conf/cvpr/SchroffKP15}, PFE \cite{DBLP:conf/iccv/ShiJ19}, ArcFace \cite{DBLP:conf/cvpr/DengGXZ19}, MagFace \cite{DBLP:conf/cvpr/MengZH021}, and QMagFace \cite{DBLP:journals/corr/abs-2111-13475}.

\subsection{Confidence Estimation Approaches}
In the experiments, we compare the proposed probabilistic-interpretable comparison (PIC) score against all the other decision confidence estimation methods for face recognition that we are aware of.
This includes score-based decision confidence methods such as the distance to (decision) threshold confidence (DTC) metric \cite{DBLP:journals/corr/abs-2210-10354} (BMVC22), the likelihood ratio-based confidence (LRC) score \cite{DBLP:conf/btas/ZeinstraMVS18} (BTAS18), and the error rate based confidence (ERBC) \cite{huberICPR22} (ICPR22).
Other approaches need probabilistic face embeddings to make use of feature uncertainties for predicting decision confidence.
This includes uncertainty-propagation for matching confidence (UPMC) \cite{DBLP:journals/corr/abs-2210-10354} (BMVC22) and the probabilistic face embedding score (PFES) \cite{DBLP:conf/iccv/ShiJ19} (ICCV19).

%%%%%%%%% Results
\section{Results}
\label{sec:Results}

\subsection{Score Distribution Analysis}
\label{sec:ScoreDistributionAnalysis}

\begin{figure*}[t]
\centering
\subfloat[Adience - Comparison Scores\label{fig:ScoreDistributionAdience}]{%
     \includegraphics[width=0.247\textwidth]{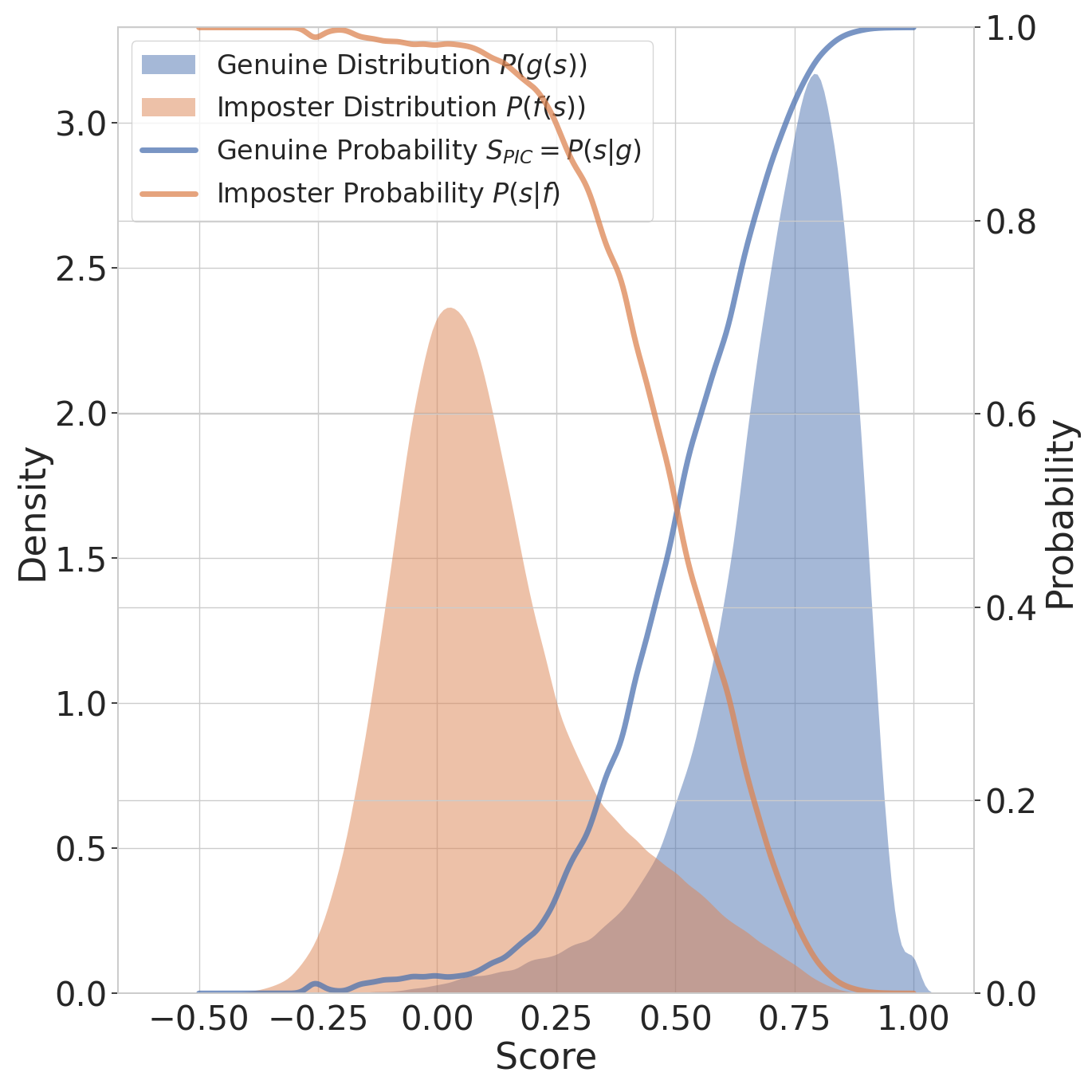}} 
\subfloat[ColorFeret - Comparison Scores\label{fig:ScoreDistributionColorFeret}]{%
     \includegraphics[width=0.247\textwidth]{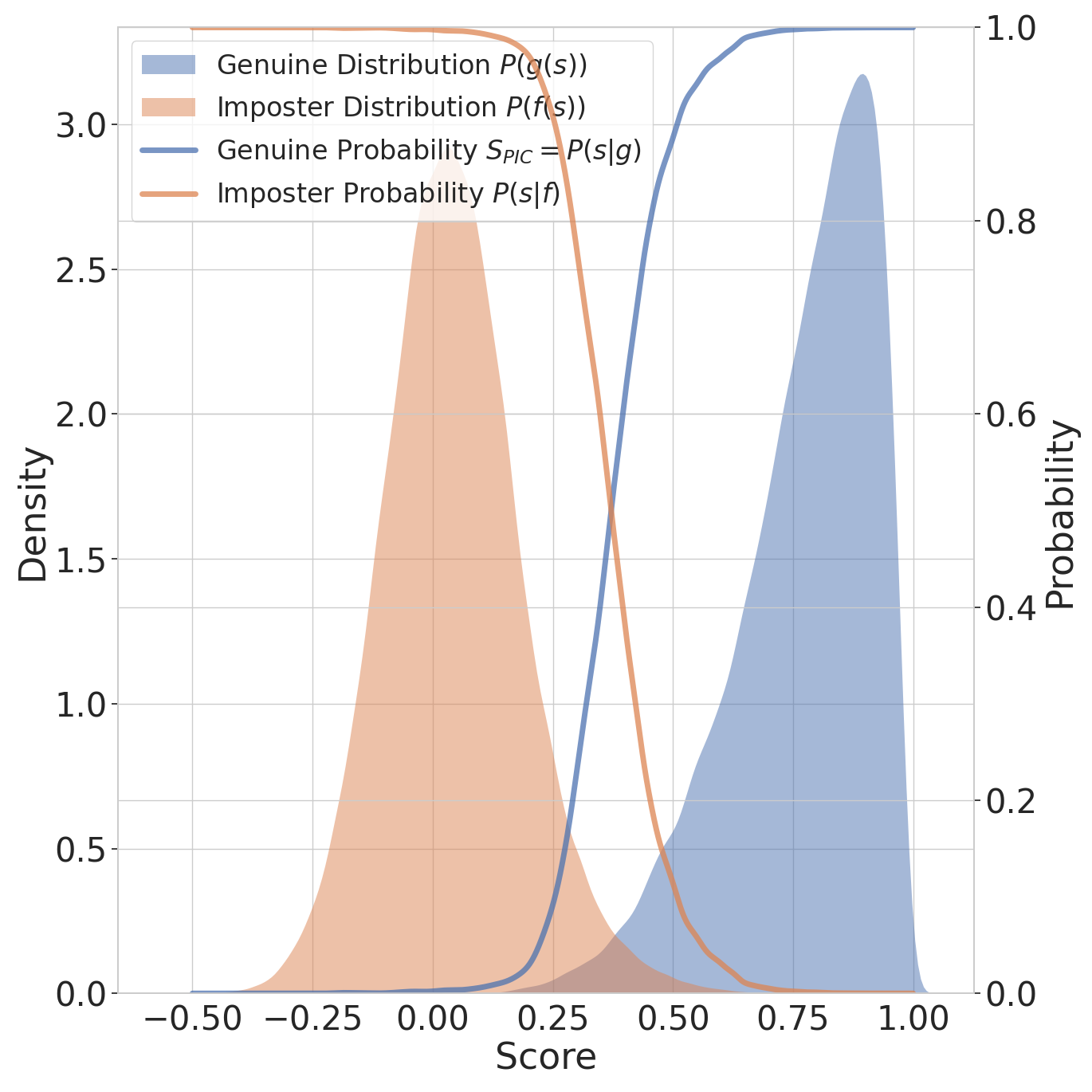}} 
\subfloat[LFW - Comparison Scores\label{fig:ScoreDistributionLFW}]{%
     \includegraphics[width=0.247\textwidth]{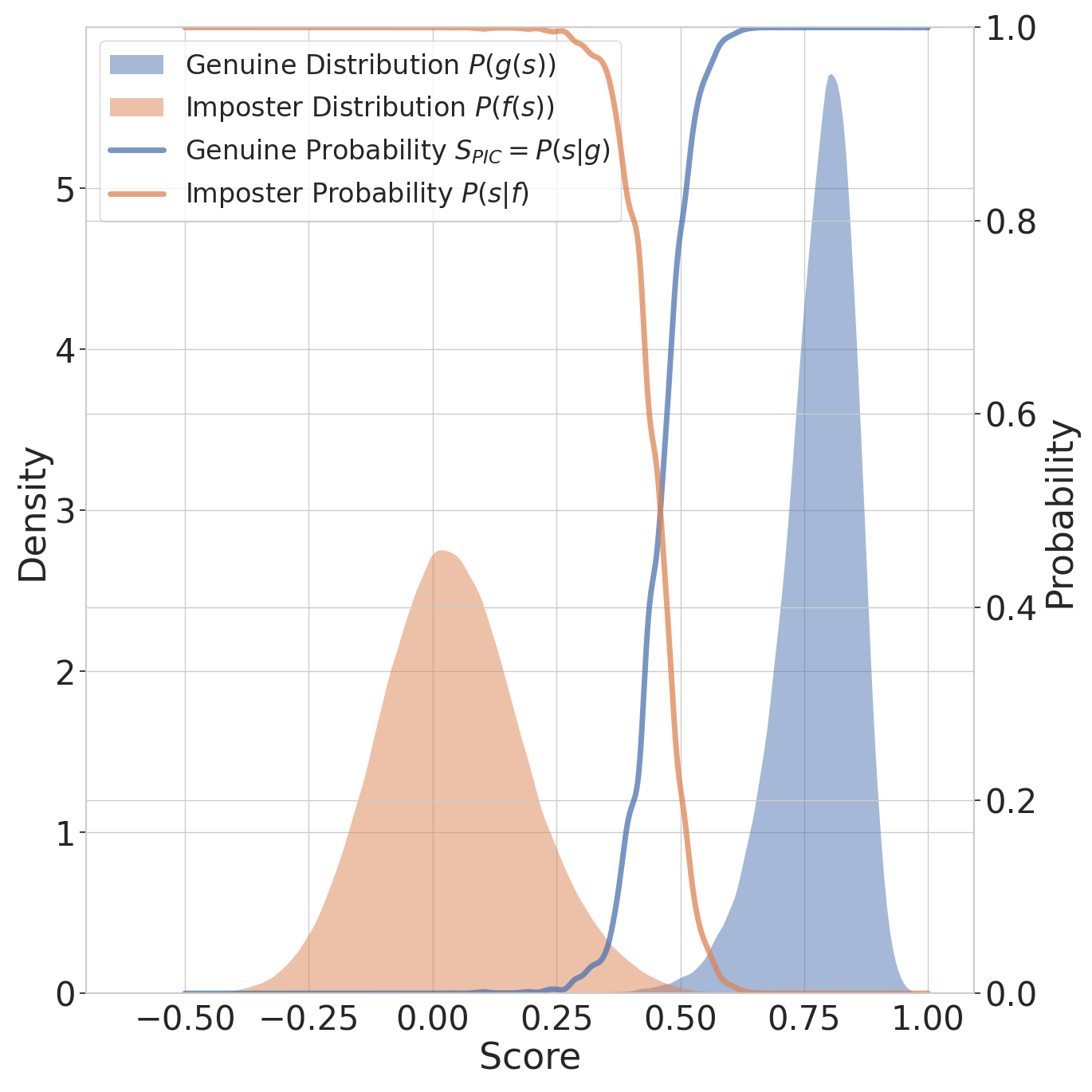}} 
\subfloat[Morph - Comparison Scores\label{fig:ScoreDistributionMorph}]{%
     \includegraphics[width=0.247\textwidth]{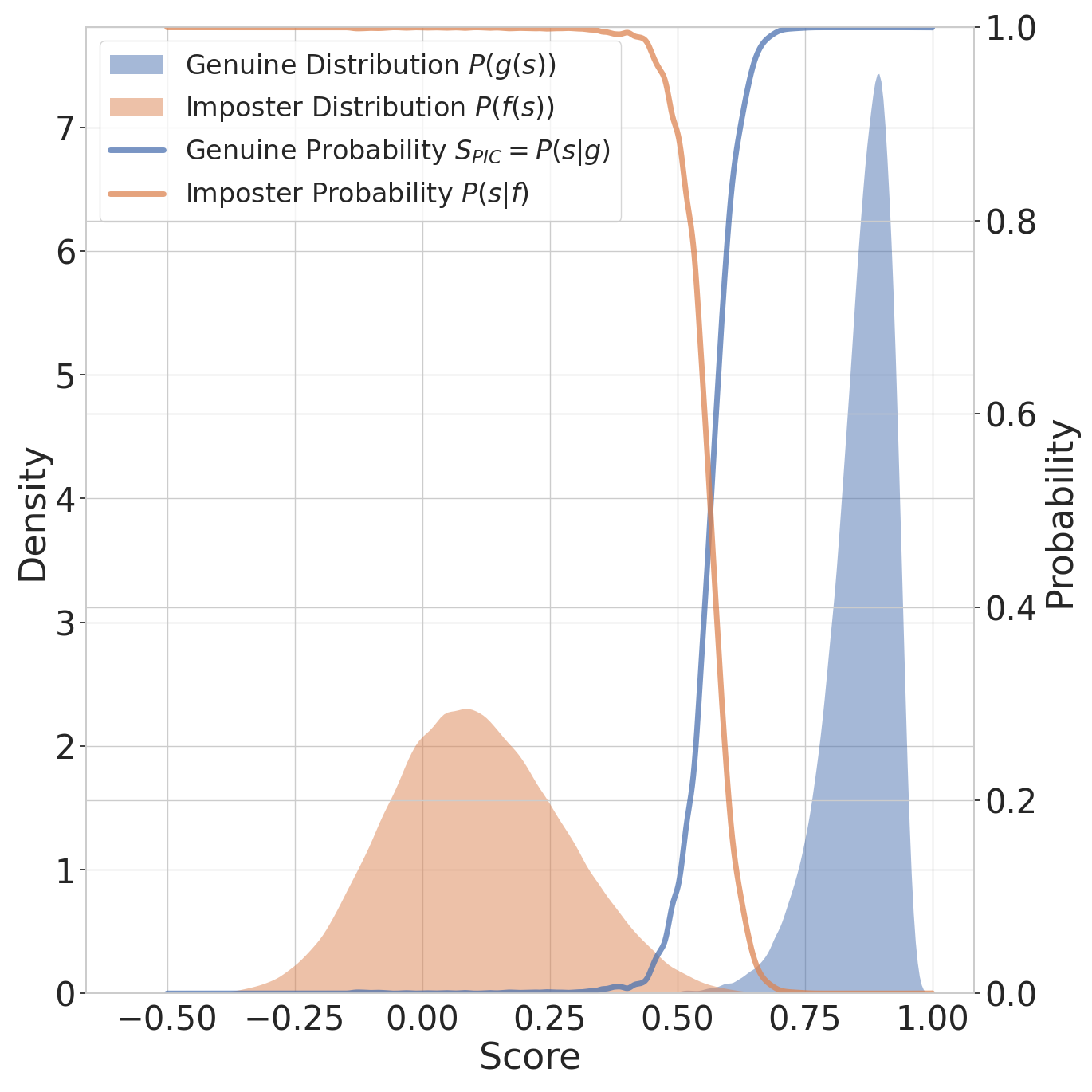}} 

\subfloat[Adience - PIC Scores\label{fig:ScoreDistributionAdience}]{%
     \includegraphics[width=0.247\textwidth]{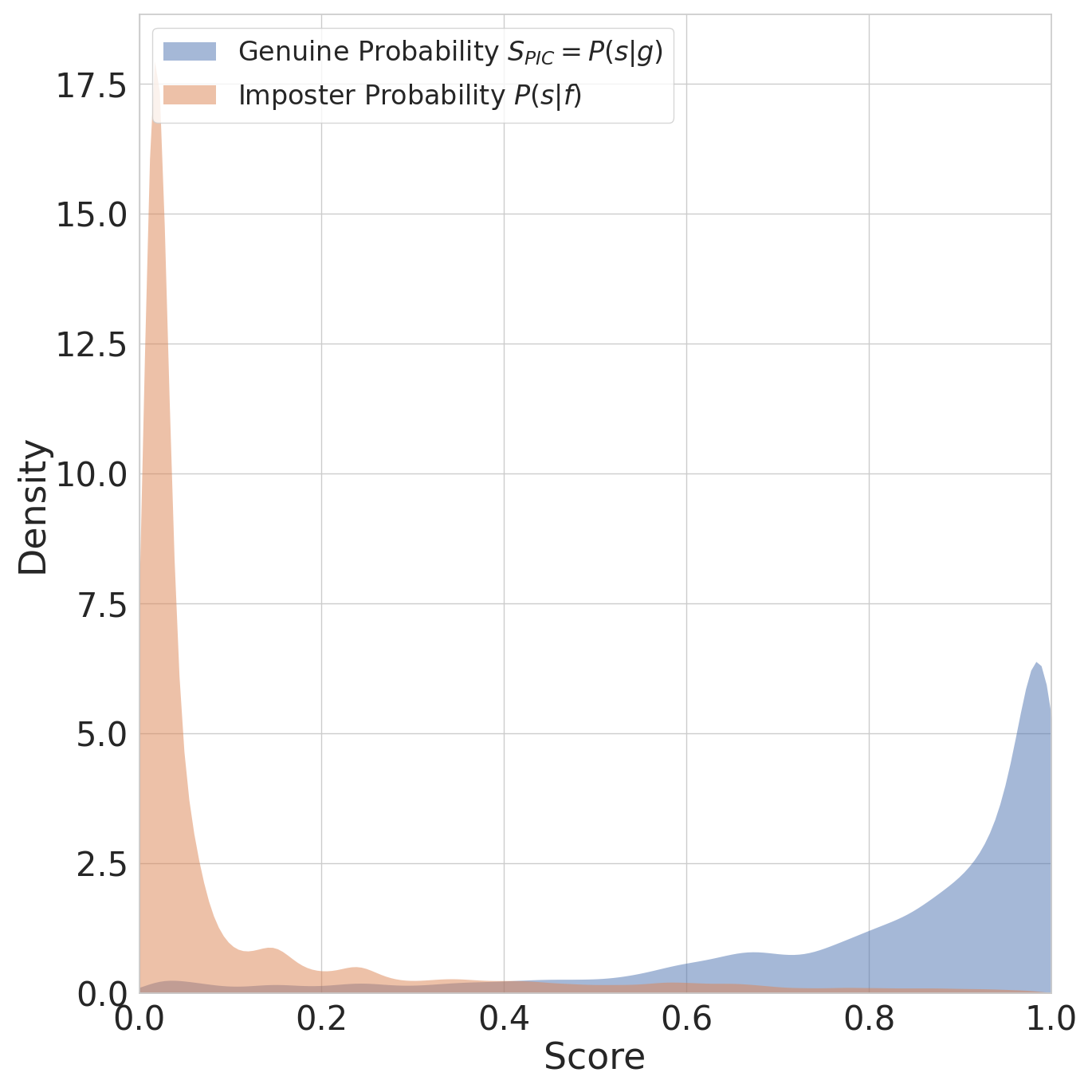}} 
\subfloat[ColorFeret - PIC Scores\label{fig:ScoreDistributionColorFeret}]{%
     \includegraphics[width=0.247\textwidth]{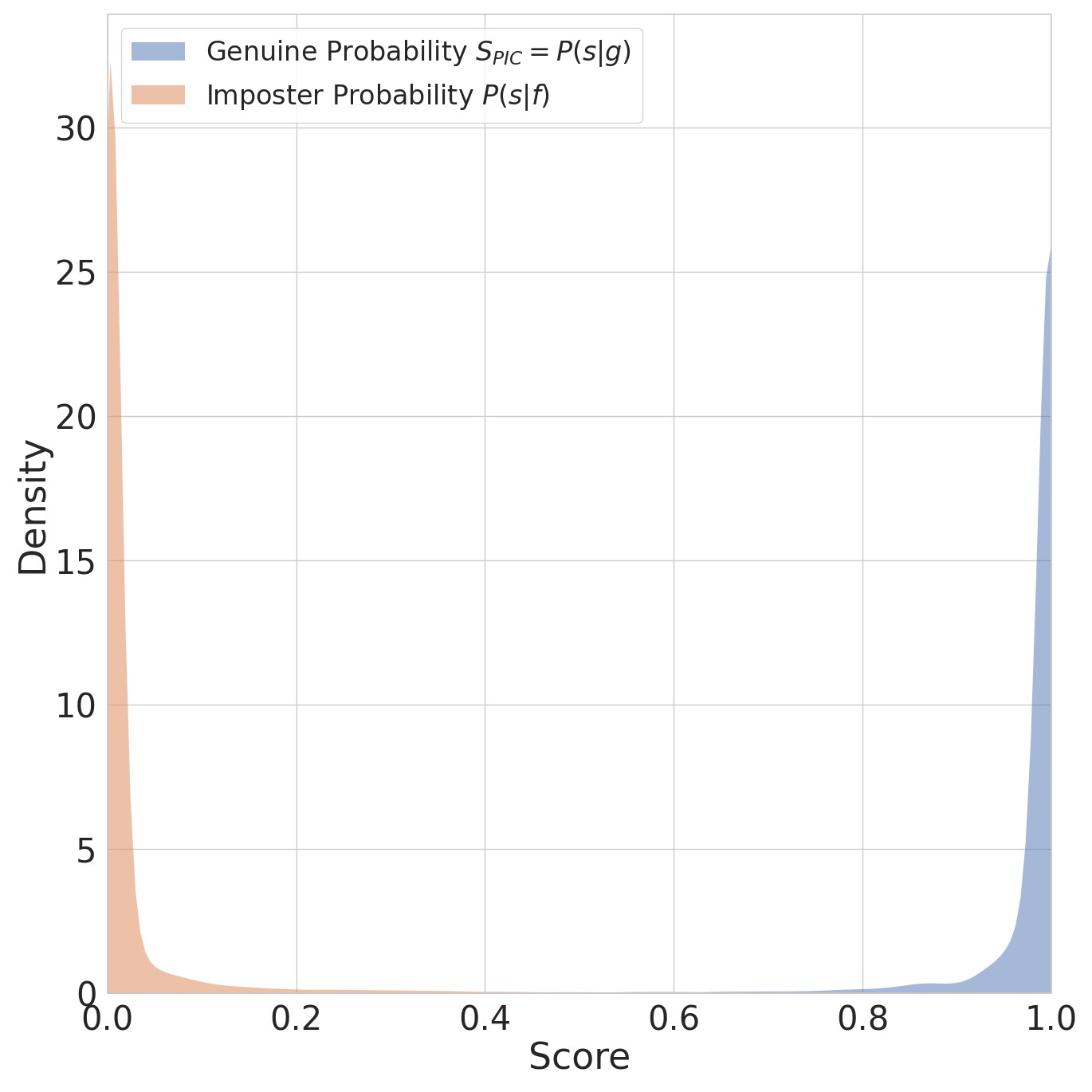}} 
\subfloat[LFW - PIC Scores\label{fig:ScoreDistributionLFW}]{%
     \includegraphics[width=0.247\textwidth]{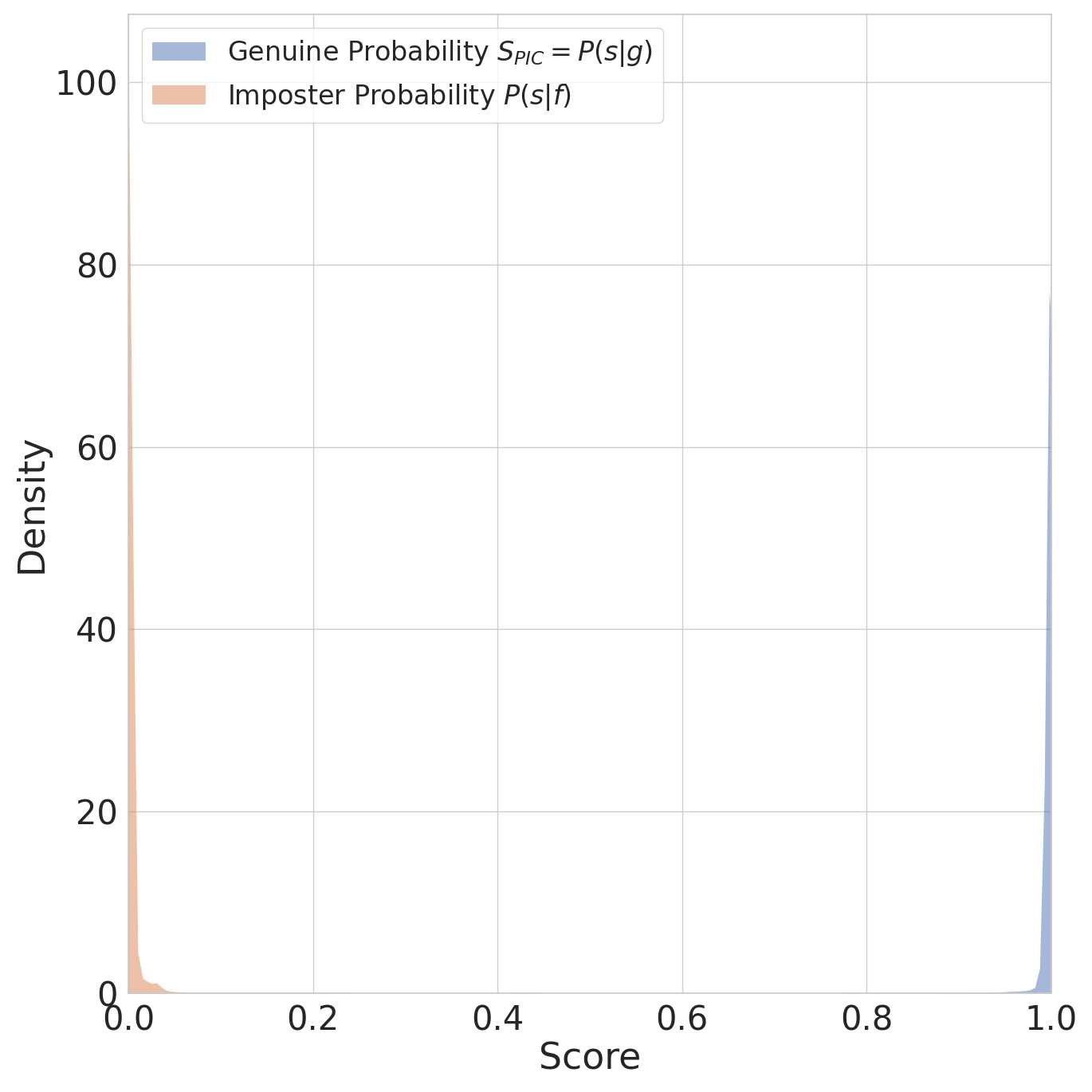}} 
\subfloat[Morph - PIC Scores\label{fig:ScoreDistributionMorph}]{%
     \includegraphics[width=0.247\textwidth]{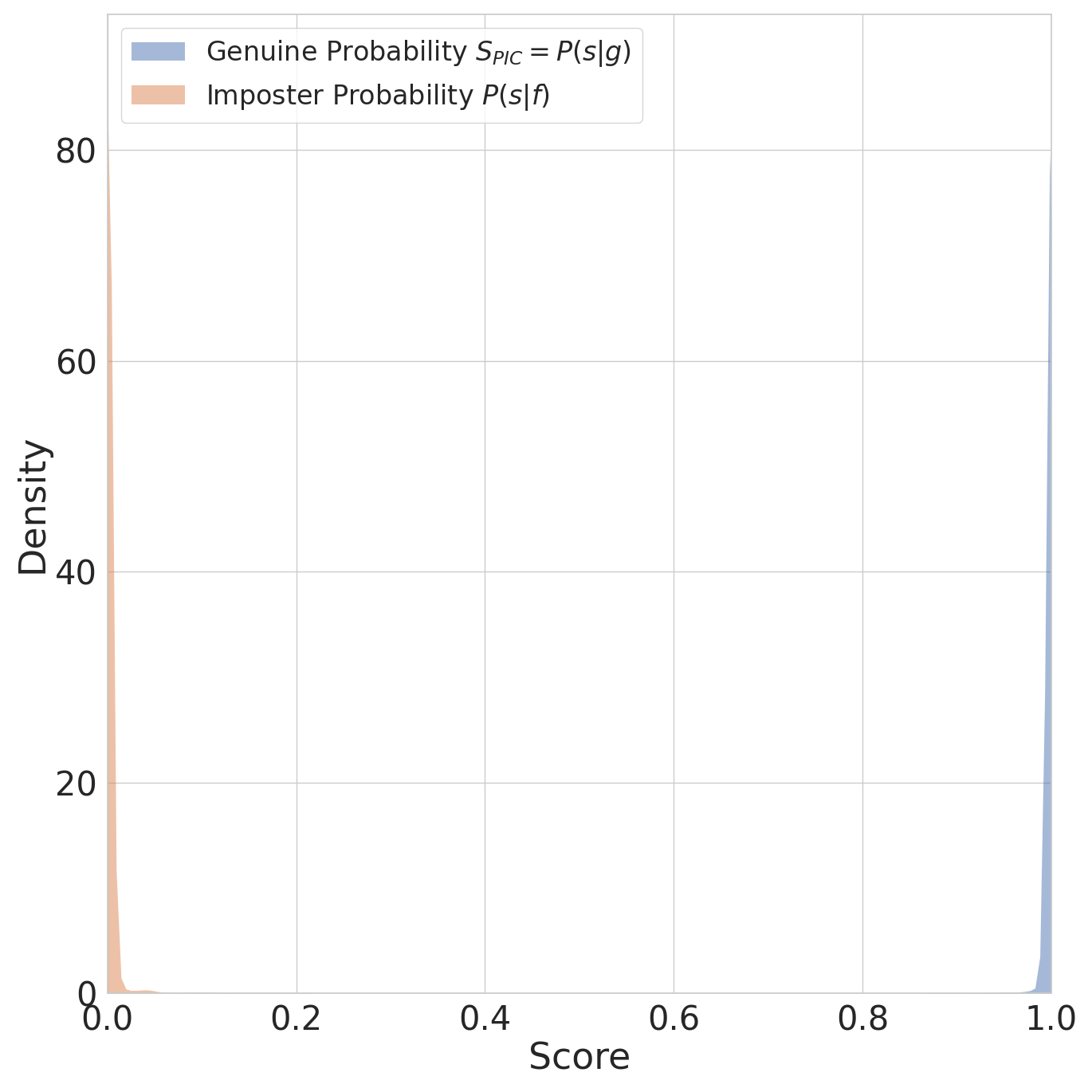}} 
\caption{ \textbf{Score Distribution Analysis} -The original (top) and the PIC (bottom) score distributions are shown for FaceNet. Based on the original score distributions, also the corresponding probabilities (see Sec. \ref{sec:Methodolgy-PIC}) for genuine and imposter are shown at the top.
On the bottom, the optimal probabilistic interpretable comparison (PIC) scores are shown. Since the PIC score distributions build on a monotonic transformation, the order of the scores, and thus the performance, remains the same. 
%For single comparisons, the PIC score only adds probabilistic interpretability.
} 
\label{fig:ScoreDistributionAnalysis}
\end{figure*}

To understand how the original comparison score is transformed to gain a probabilistic interpretation, Figure \ref{fig:ScoreDistributionAnalysis} shows the original genuine and imposter score distributions for FaceNet\footnote{Since the distributions for the other FRS lead to similar observations, we refer to the supplementary.}, as well as the corresponding PIC score distributions.
In the top row, the original score distributions for the different datasets are shown.
For the Adience and ColorFeret datasets, the genuine and imposter score distributions strongly overlap due to the challenges of these datasets, such as low image quality and strong pose differences.
LFW and Morph show mostly frontal and well-illuminated faces and thus, the score distributions show significantly less overlap. 
Consequently, on these "easy" datasets, LFW and Morph, fewer samples for the probabilities in the overlap areas exist.
This will lead to some unstable results in the confidence calibration analysis, as we will see in Section \ref{sec:SingeComparisonCalibrationAnalysis}.
Additionally to the score distributions, the optimal probability that a sample with a score of $s$ belongs to genuine or imposter (see Sec. \ref{sec:Methodolgy-PIC}) is shown. 
Intuitively, both probabilities equalize when the number of genuine and imposter scores for $s$ are the same.

In the bottom row, the score distributions for the proposed PIC scores are shown.
The score refers to the probability that a sample belongs to a genuine comparison.
Since the order of the scores remains the same for the original and the PIC scores, this will lead to similar single-biometric recognition performances.
For a single sample, the PIC score adds the optimal probabilistic interpretability for a generic biometric system.
However, contrary to the standard comparison score, a PIC score for multiple comparisons can be calculated as demonstrated in Section \ref{sec:MultiComparisonAnalysis}.

%Score distribution

\subsection{Single-Comparison Calibration Analysis}
\label{sec:SingeComparisonCalibrationAnalysis}

To analyze the probabilistic interpretability of different confidence estimation approaches, we introduce confidence calibration curves (CCC).
A CCC compares the true confidence of each sample (x-axis) in a given test set bin-wise with the average predicted confidence (y-axis).
%of different approaches.
For a perfectly calibrated confidence estimator, the CCC shows a linear bisectrix line.
To compute a CCC, the true confidence of each test set sample is calculated and divided into $b=30$ bins.
For each bin, the mean and standard deviation of the predicted confidences are computed.
Please note that in the LFW or Morph datasets, the number of samples for single bins might be low.
Since these datasets are less challenging (see Section \ref{sec:ScoreDistributionAnalysis}) and thus, provide fewer samples for specific probabilities, the performance becomes more unstable in these cases.

\begin{figure*}[t]
\centering
\subfloat[Adience - FaceNet\label{fig:ScoreDistributionAdience}]{%
     \includegraphics[width=0.21\textwidth]{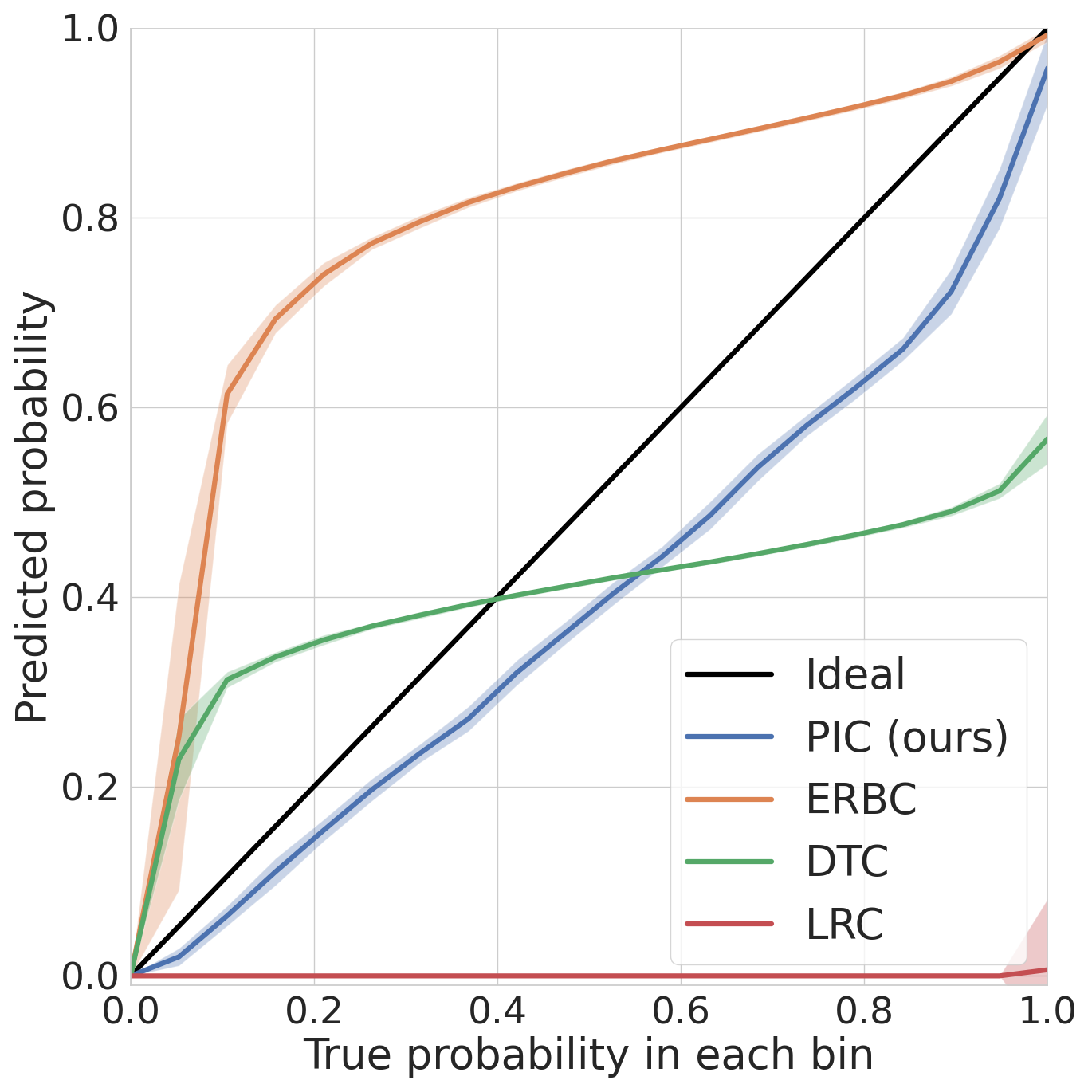}} 
\subfloat[ColorFeret - FaceNet\label{fig:ScoreDistributionColorFeret}]{%
     \includegraphics[width=0.21\textwidth]{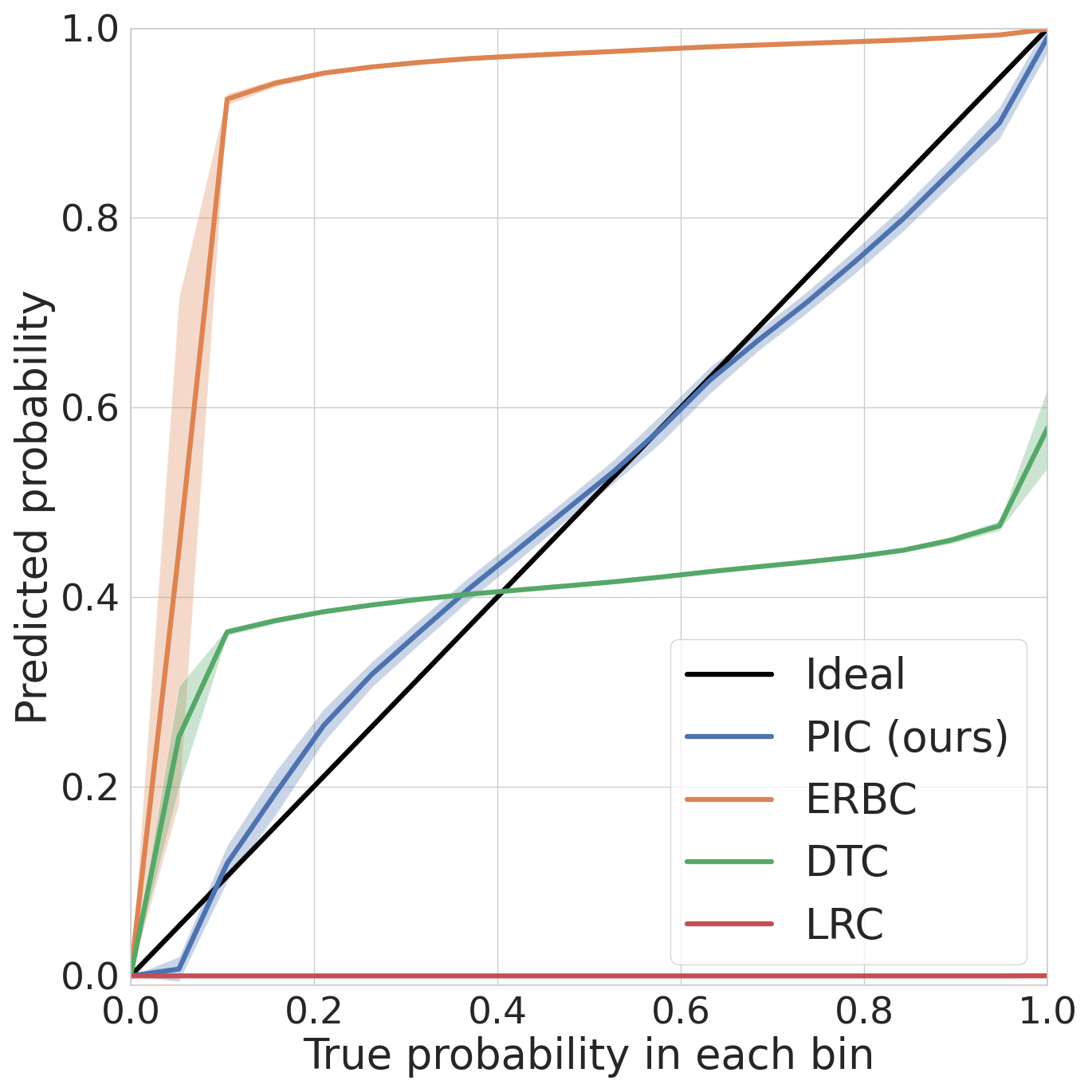}} 
\subfloat[LFW - FaceNet\label{fig:ScoreDistributionLFW}]{%
     \includegraphics[width=0.21\textwidth]{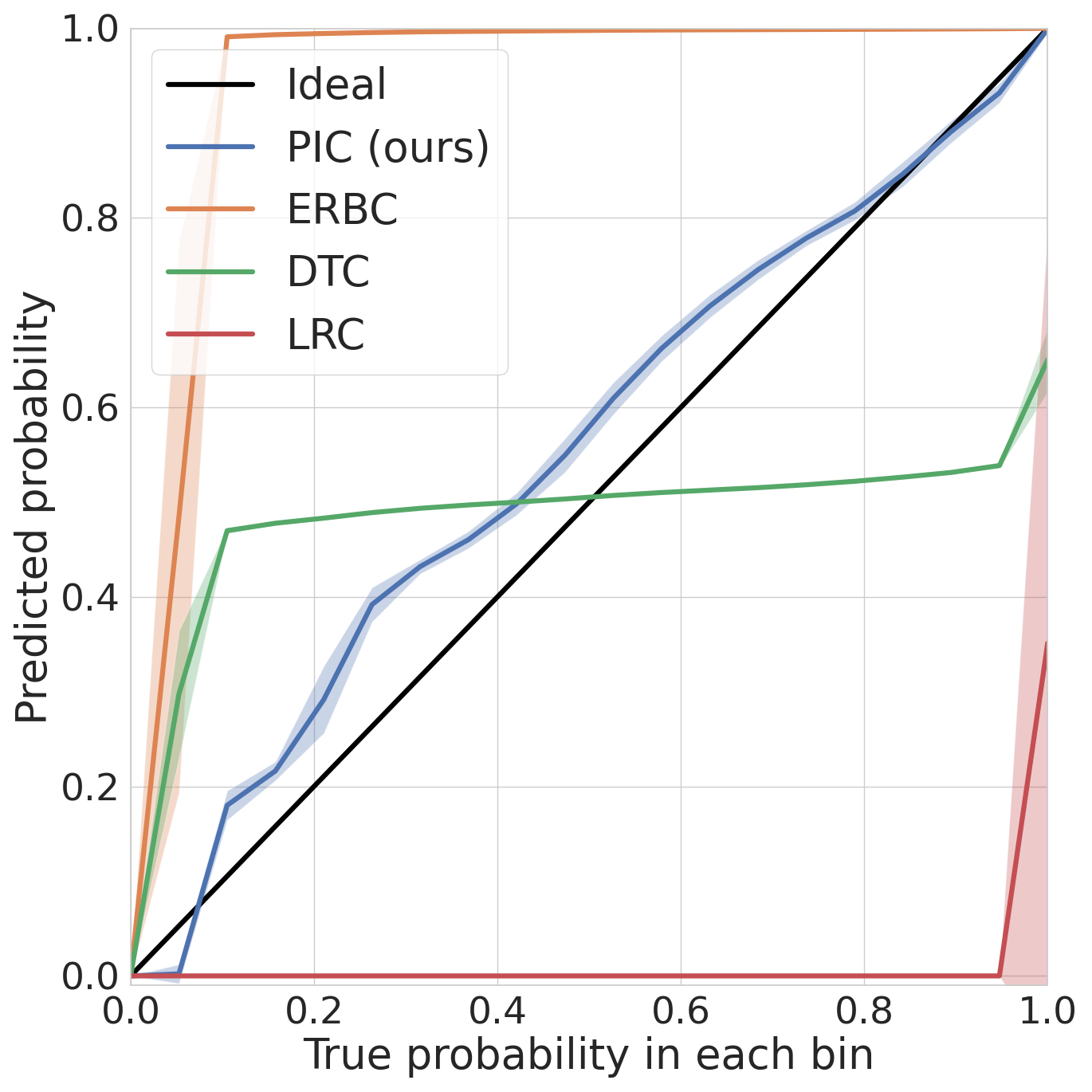}} 
\subfloat[Morph - FaceNet\label{fig:ScoreDistributionMorph}]{%
     \includegraphics[width=0.21\textwidth]{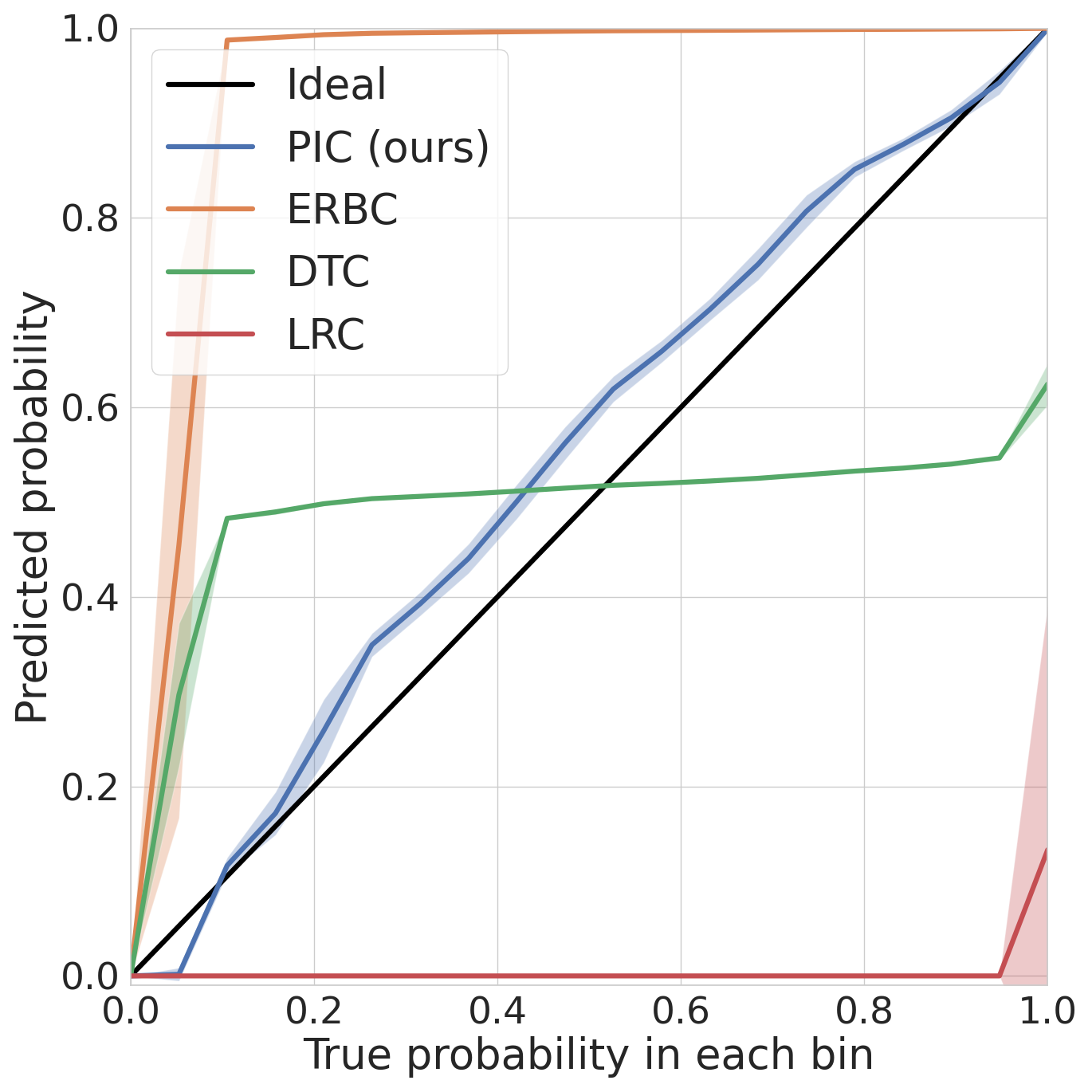}}  

\subfloat[Adience - ArcFace\label{fig:ScoreDistributionAdience}]{%
     \includegraphics[width=0.21\textwidth]{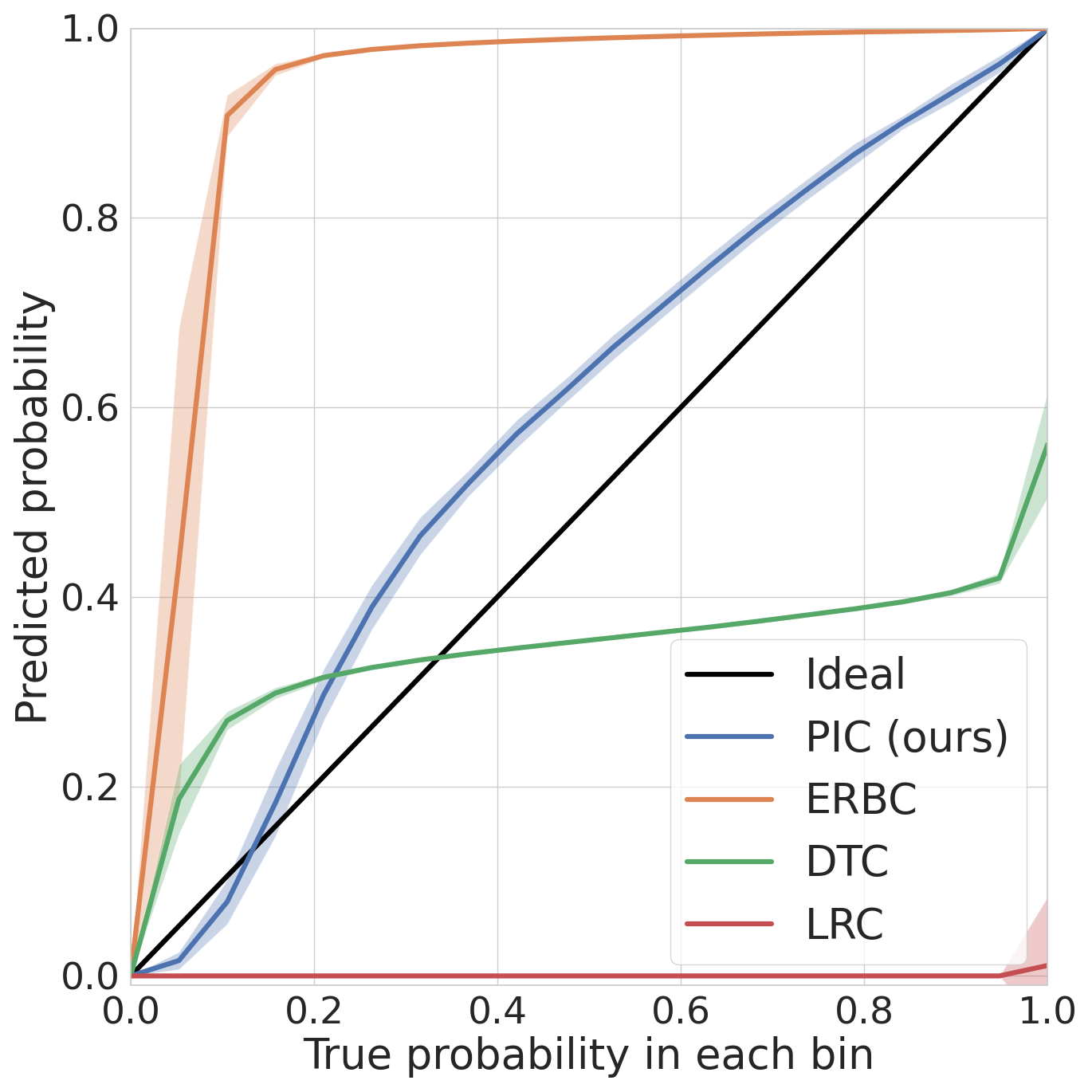}} 
\subfloat[ColorFeret - ArcFace\label{fig:ScoreDistributionColorFeret}]{%
     \includegraphics[width=0.21\textwidth]{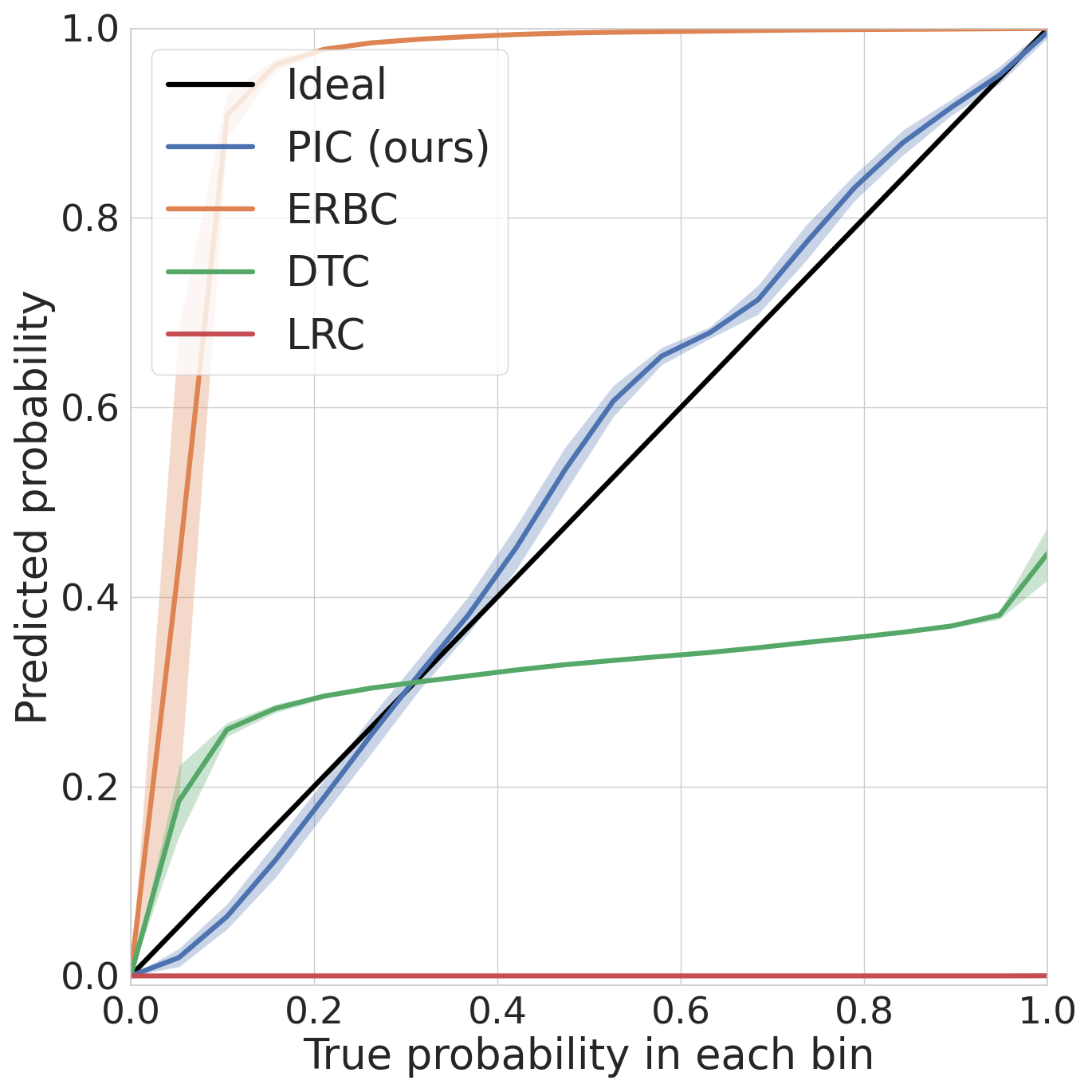}} 
\subfloat[LFW - ArcFace\label{fig:ScoreDistributionLFW}]{%
     \includegraphics[width=0.21\textwidth]{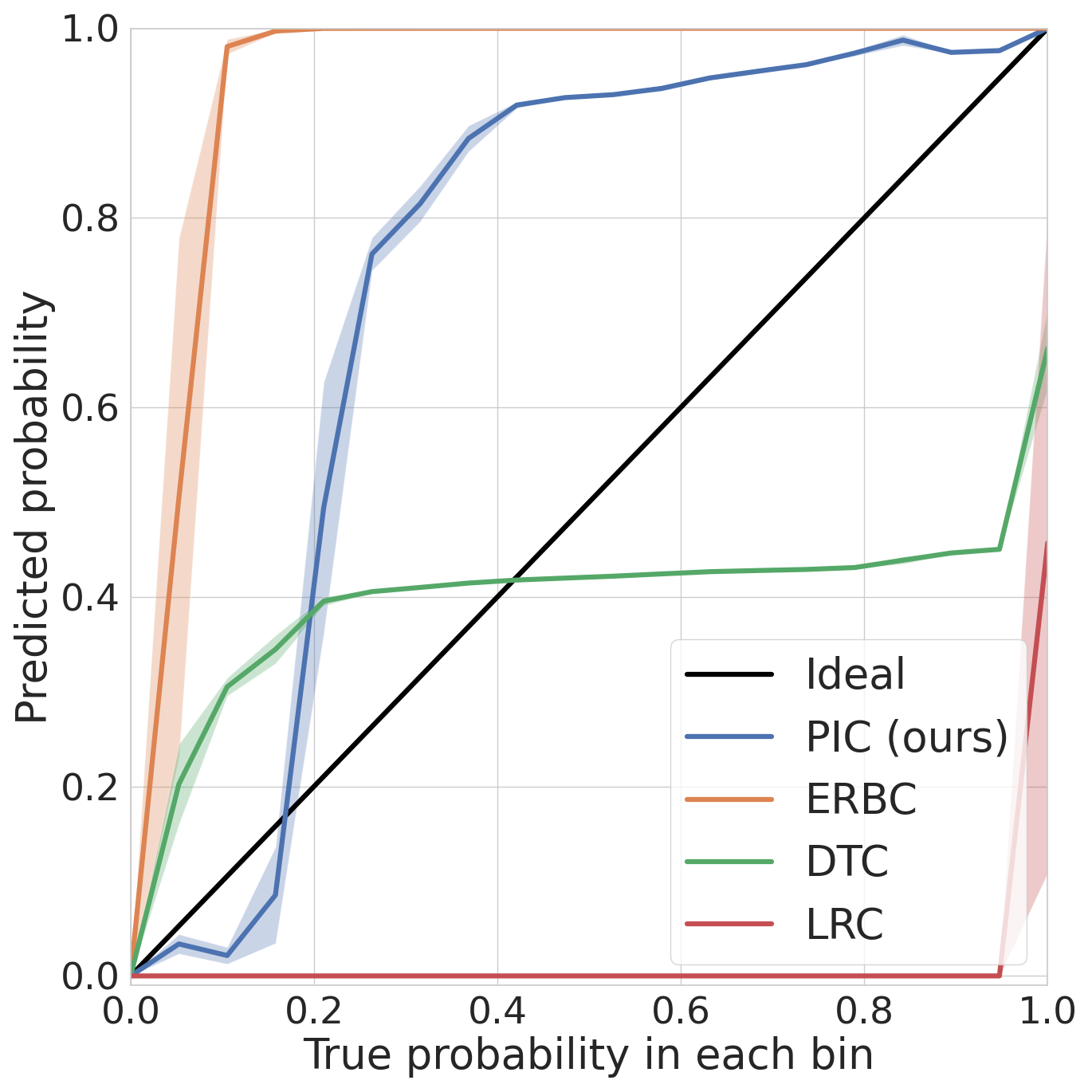}} 
\subfloat[Morph - ArcFace\label{fig:ScoreDistributionMorph}]{%
     \includegraphics[width=0.21\textwidth]{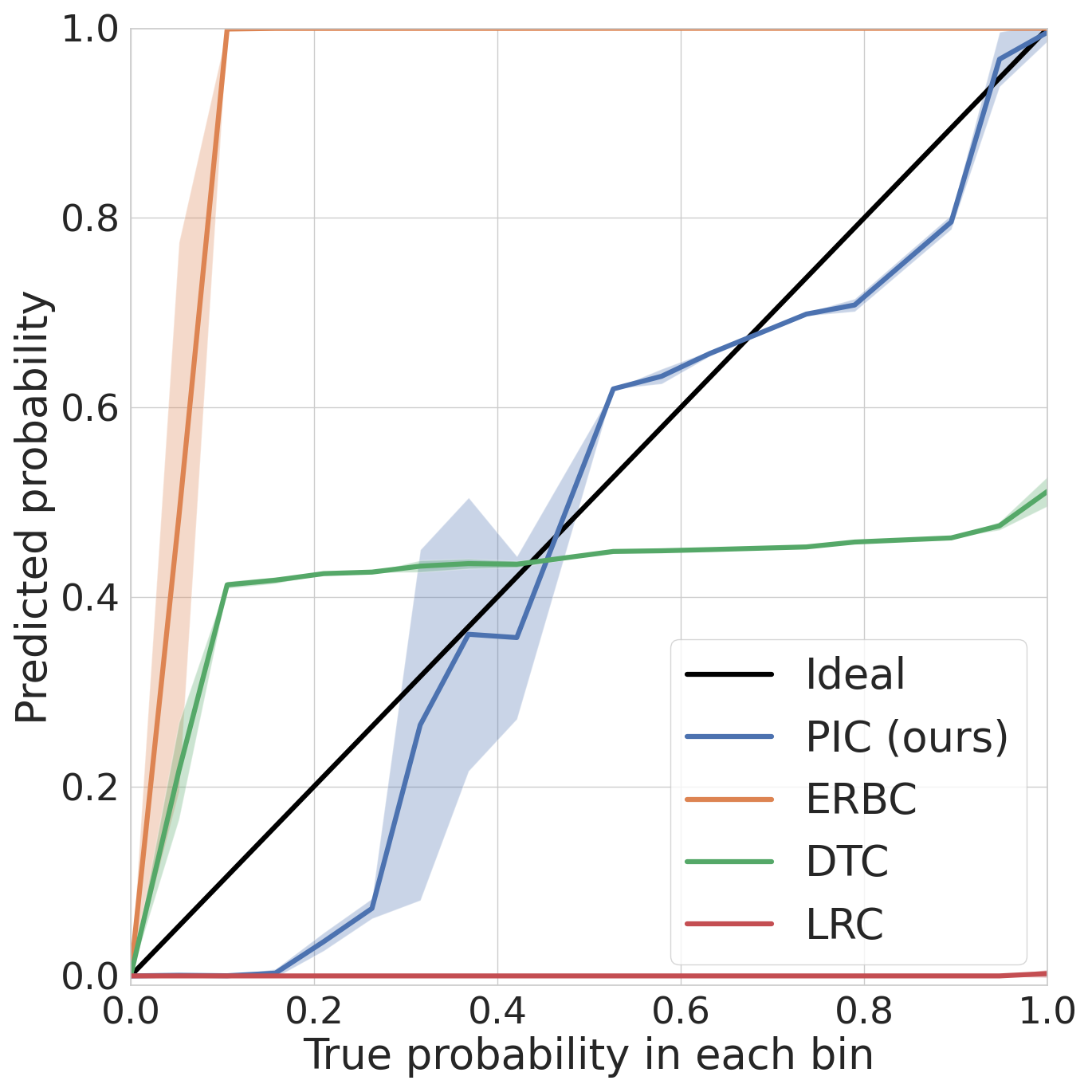}}   

\subfloat[Adience - MagFace\label{fig:ScoreDistributionAdience}]{%
     \includegraphics[width=0.21\textwidth]{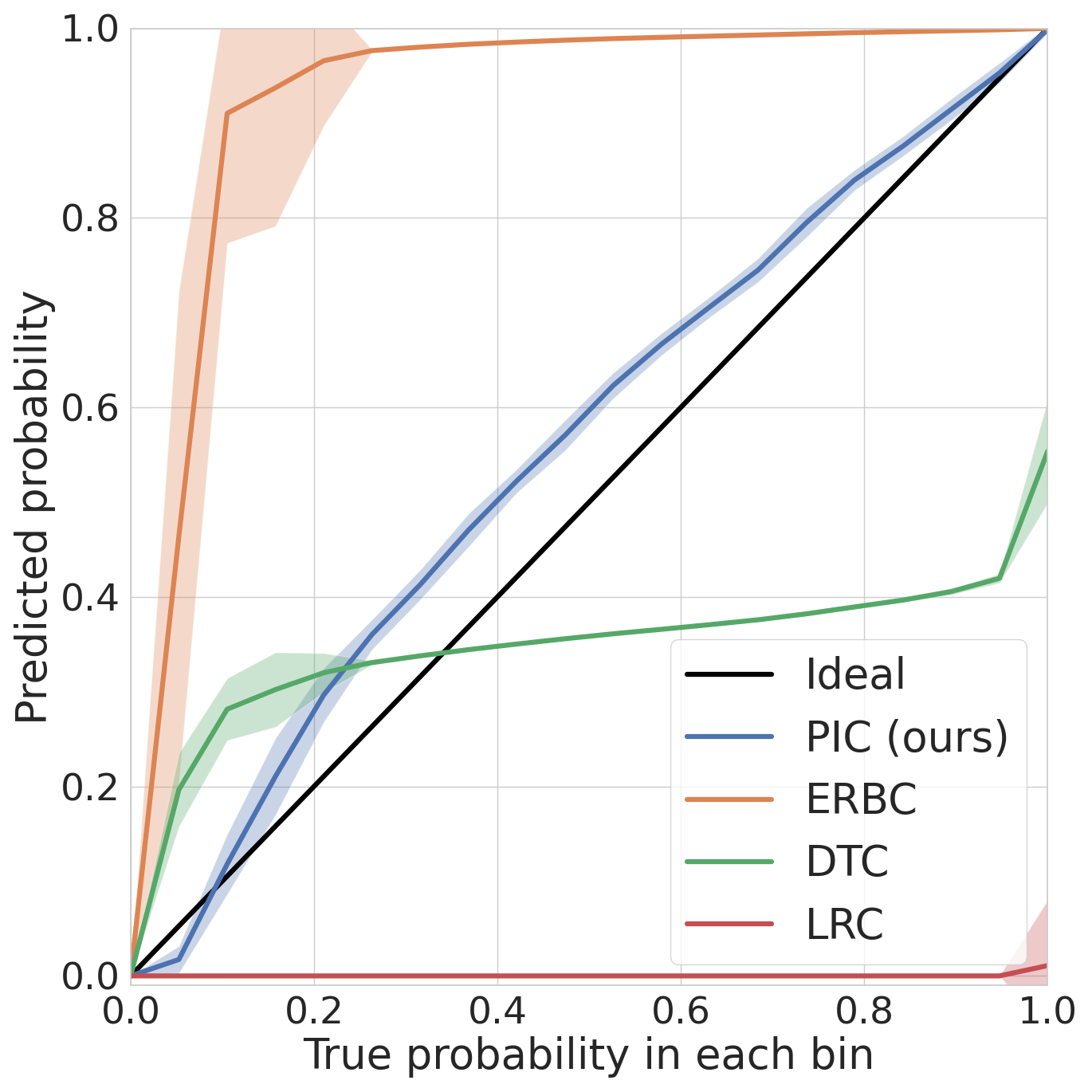}} 
\subfloat[ColorFeret - MagFace\label{fig:ScoreDistributionColorFeret}]{%
     \includegraphics[width=0.21\textwidth]{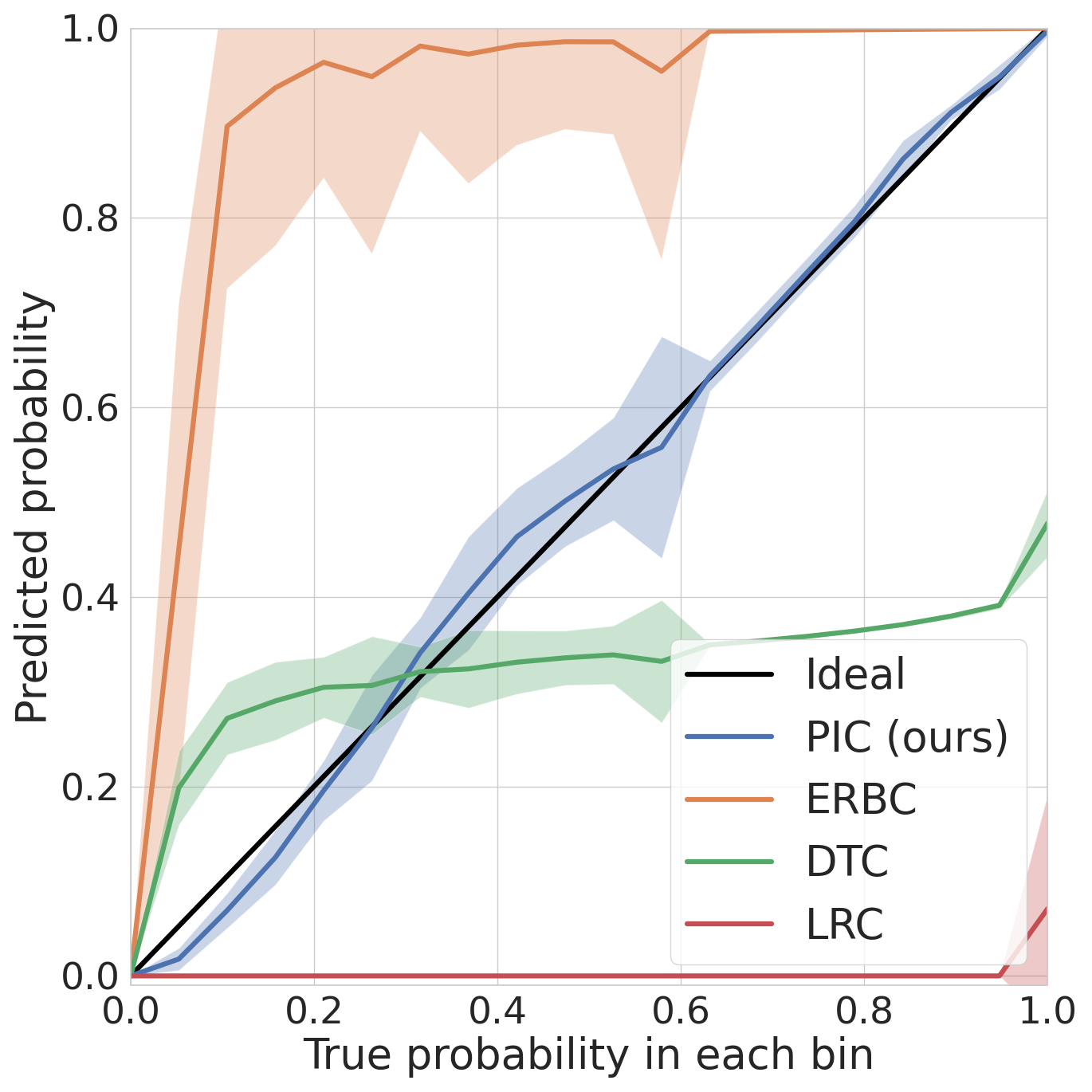}} 
\subfloat[LFW - MagFace\label{fig:ScoreDistributionLFW}]{%
     \includegraphics[width=0.21\textwidth]{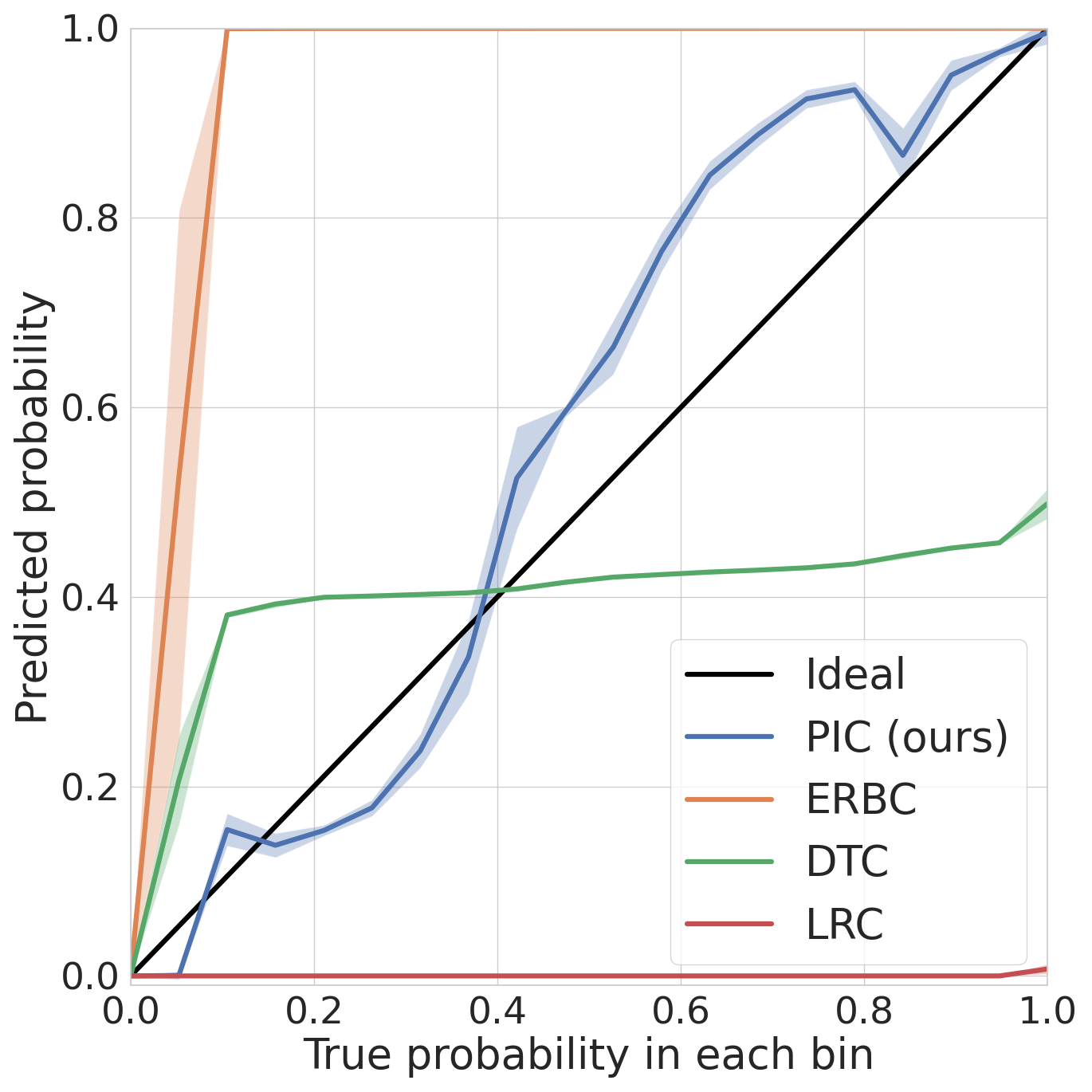}} 
\subfloat[Morph - MagFace\label{fig:ScoreDistributionMorph}]{%
     \includegraphics[width=0.21\textwidth]{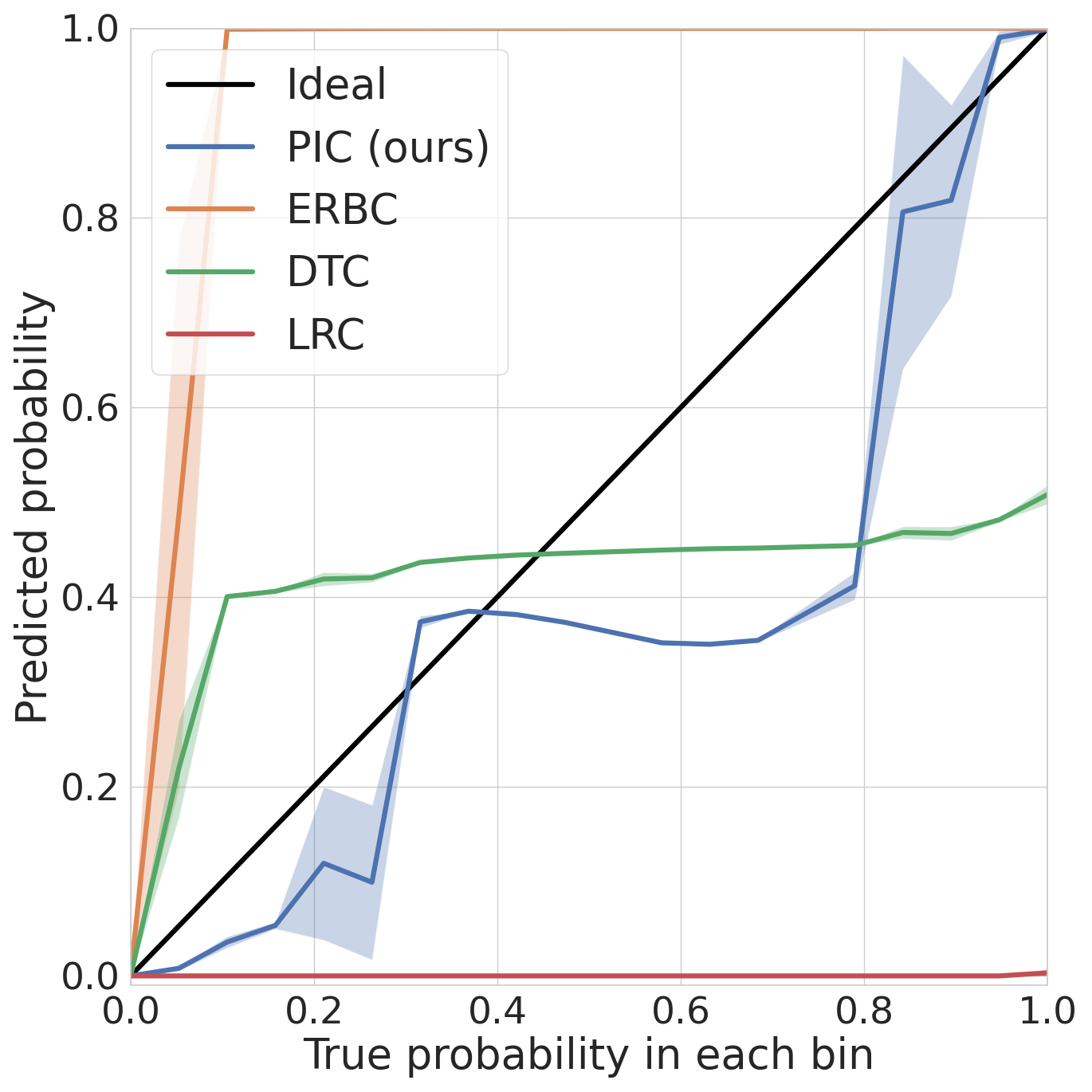}}   

\subfloat[Adience - QMagFace\label{fig:ScoreDistributionAdience}]{%
     \includegraphics[width=0.21\textwidth]{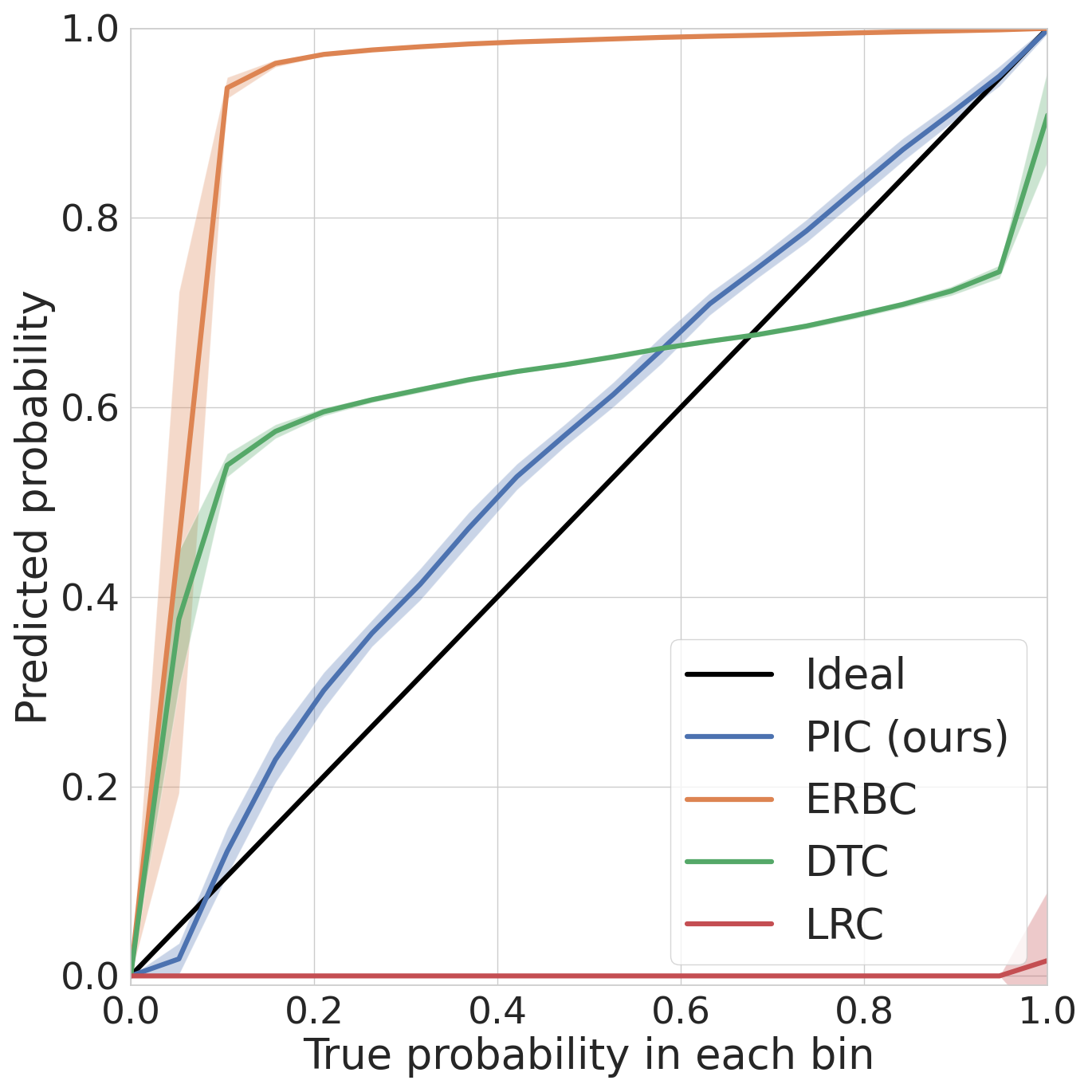}} 
\subfloat[ColorFeret - QMagFace\label{fig:ScoreDistributionColorFeret}]{%
     \includegraphics[width=0.21\textwidth]{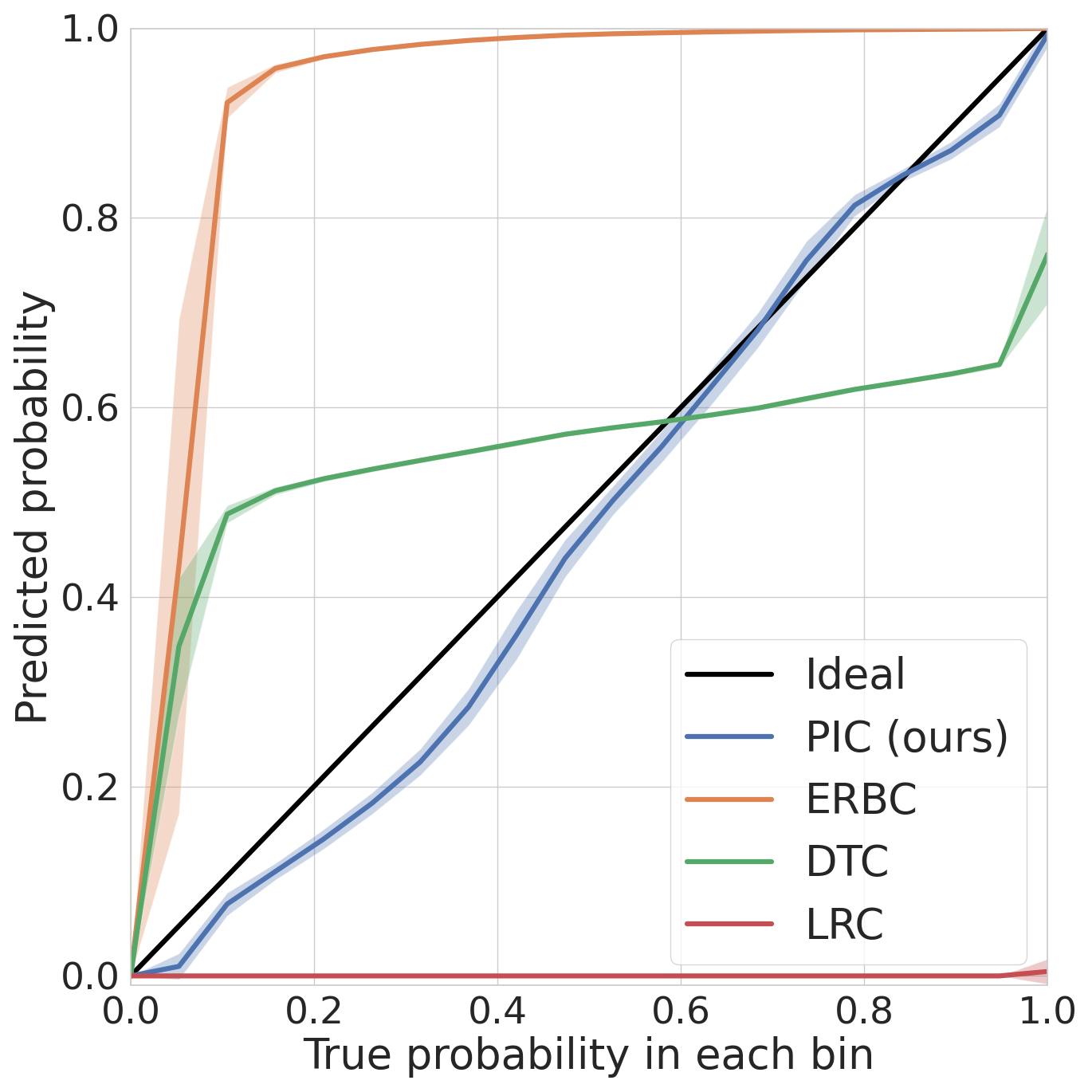}} 
\subfloat[LFW - QMagFace\label{fig:ScoreDistributionLFW}]{%
     \includegraphics[width=0.21\textwidth]{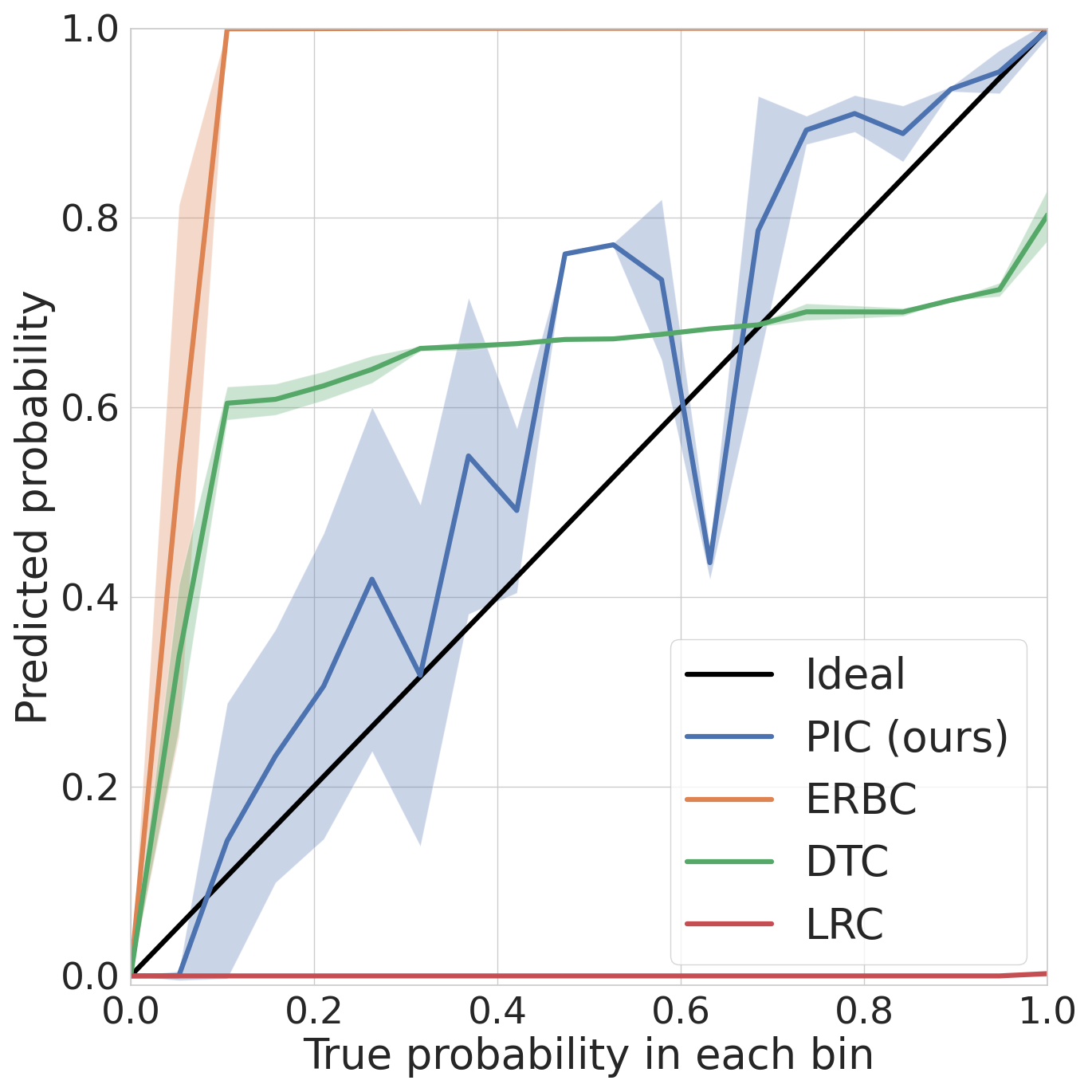}} 
\subfloat[Morph - QMagFace\label{fig:ScoreDistributionMorph}]{%
     \includegraphics[width=0.21\textwidth]{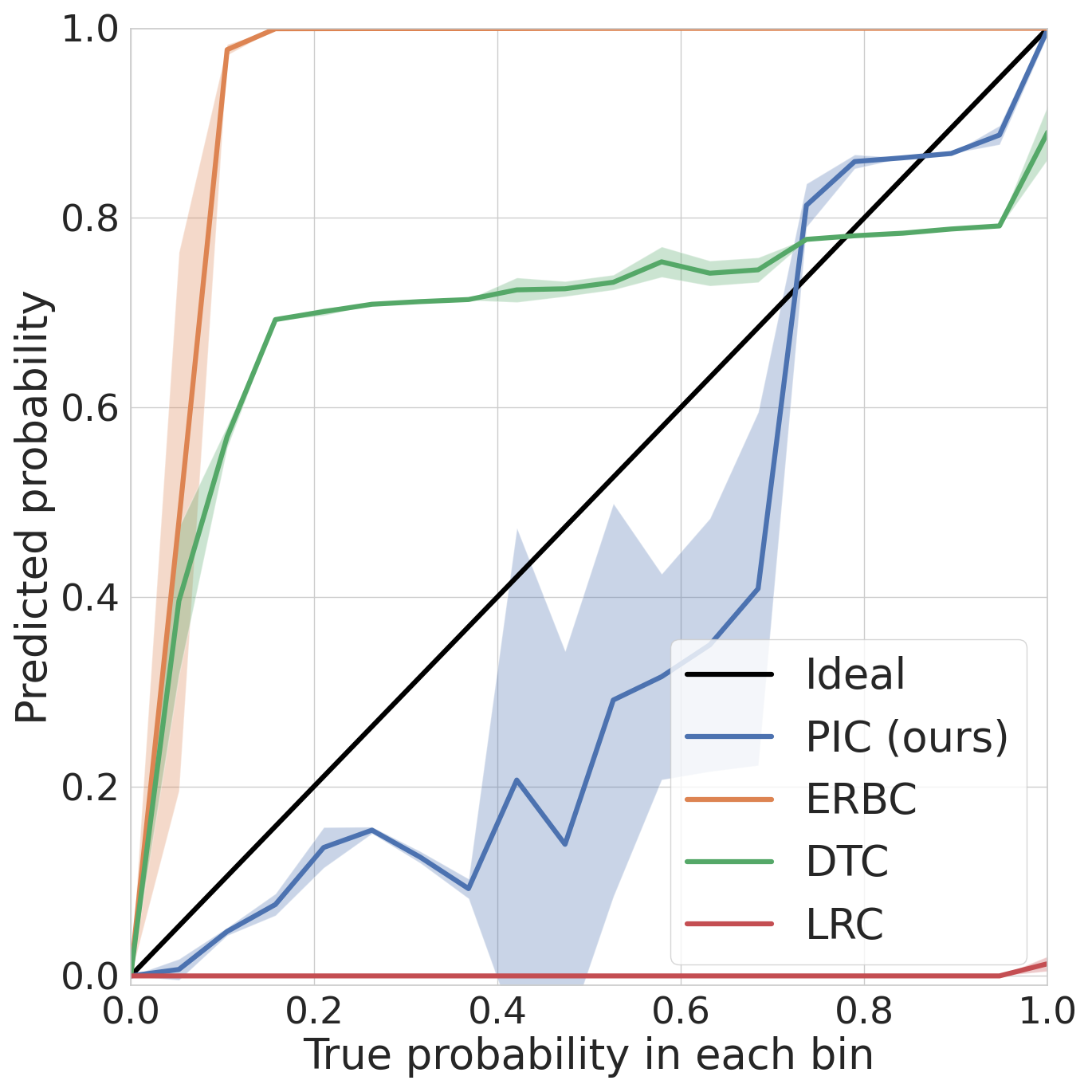}}  

\subfloat[Adience - PFE\label{fig:ScoreDistributionAdience}]{%
     \includegraphics[width=0.21\textwidth]{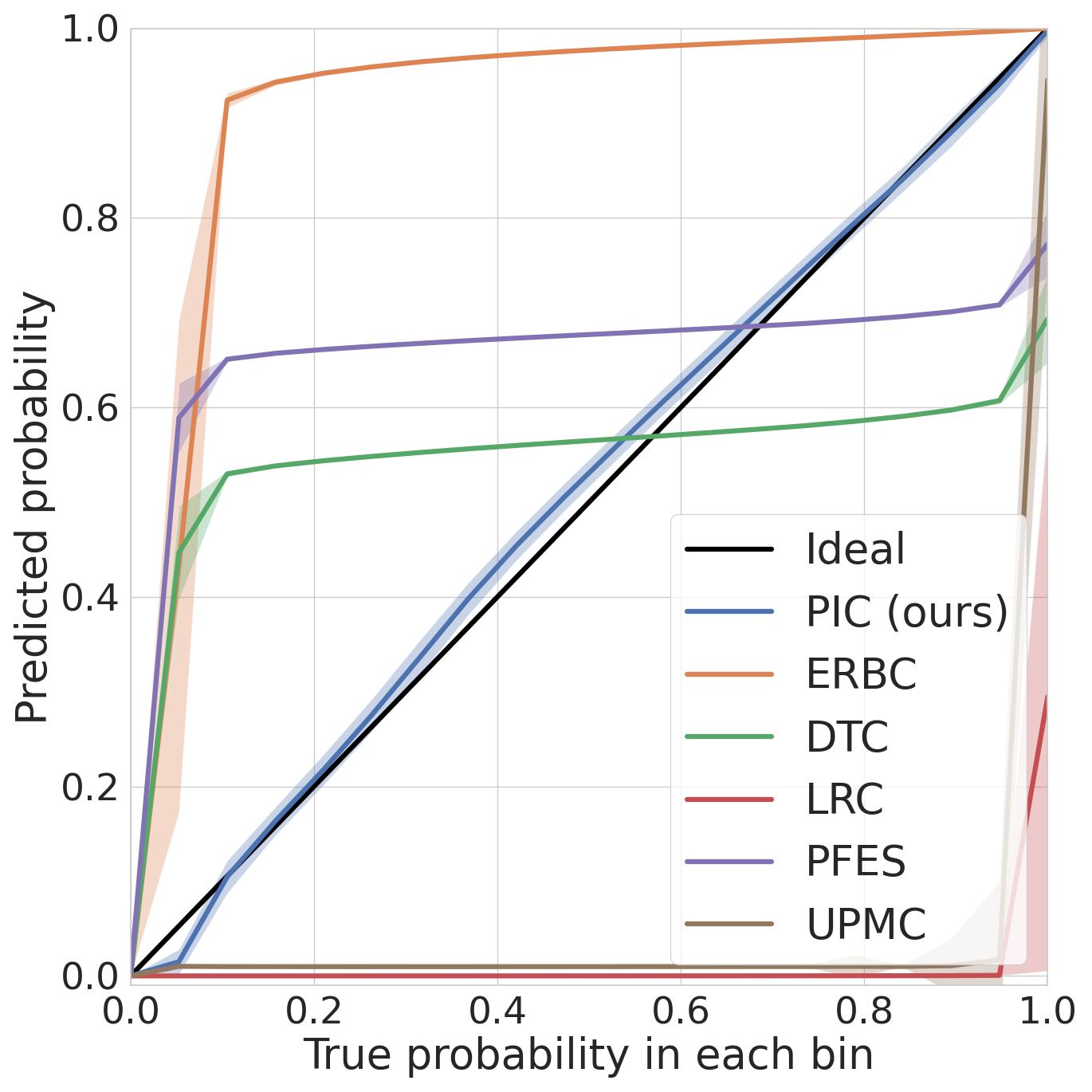}} 
\subfloat[ColorFeret - PFE\label{fig:ScoreDistributionColorFeret}]{%
     \includegraphics[width=0.21\textwidth]{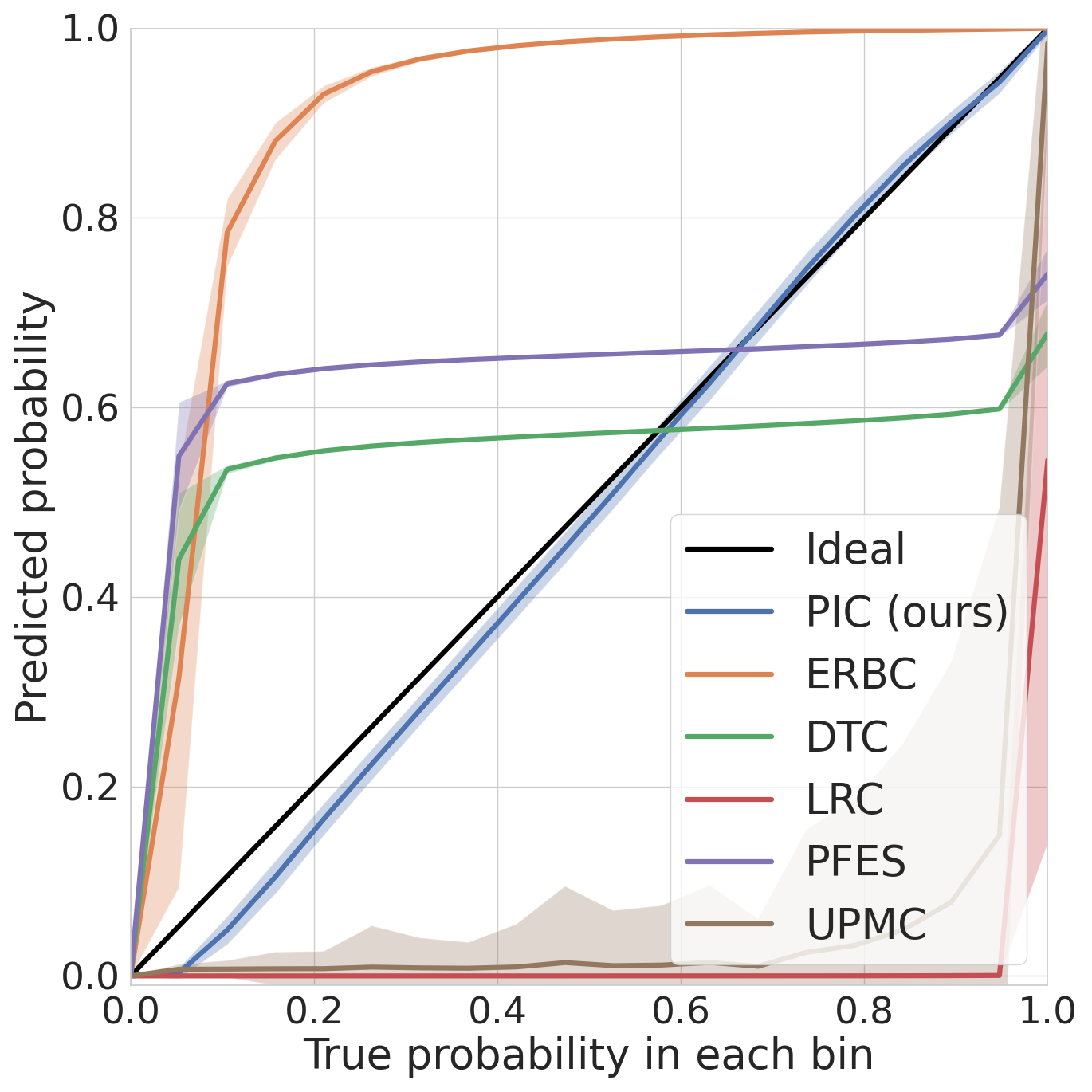}} 
\subfloat[LFW - PFE\label{fig:ScoreDistributionLFW}]{%
     \includegraphics[width=0.21\textwidth]{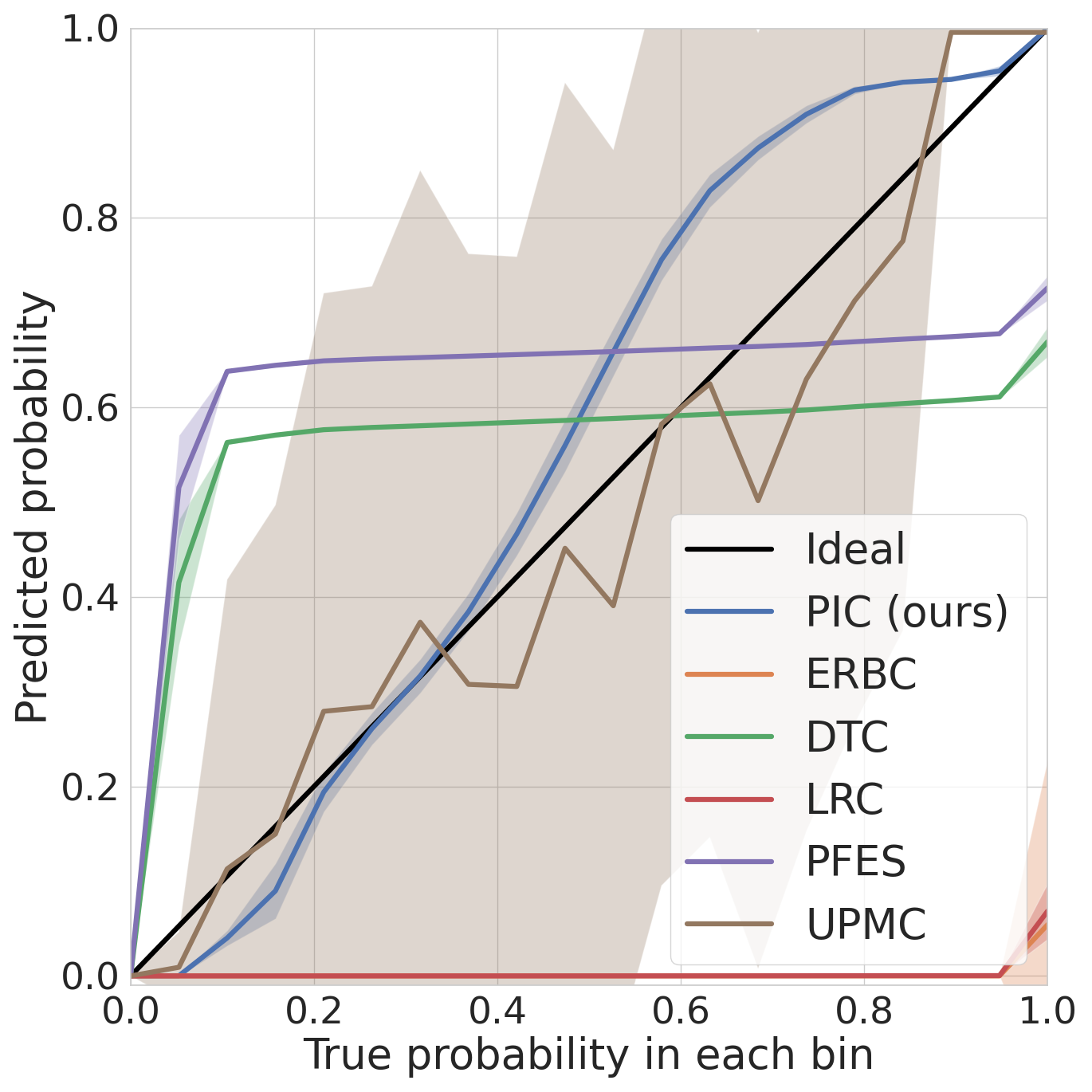}} 
\subfloat[Morph - PFE\label{fig:ScoreDistributionMorph}]{%
     \includegraphics[width=0.21\textwidth]{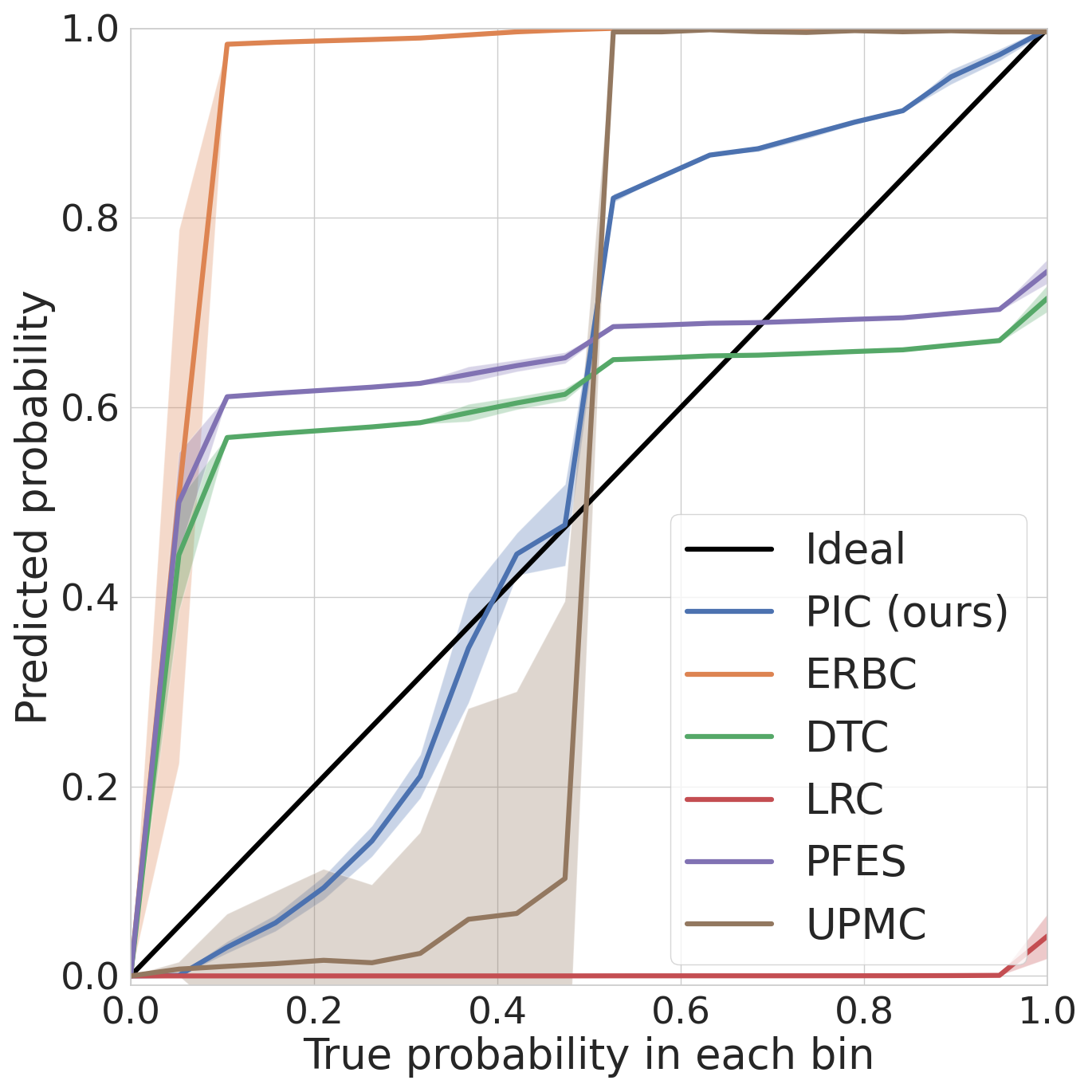}} 
\caption{\textbf{Confidence Calibration Curves (CCC)} - The CCC for all dataset and FRS combinations are shown. Inconsistencies are due to the low number of samples for specific probability bins (e.g. for LFW and Morph).
While most approaches have to deal with high under- and over-confident predictions, the proposed PIC produces close-to-ideal (black line) probabilistic confidence estimates in most cases.
} 
\label{fig:CCCplots}
\end{figure*}

Figure \ref{fig:CCCplots} shows the CCC plots for all datasets and face recognition system combinations at an FMR of $10^{-3}$.
The ideal case with an optimal confidence estimation is shown as a black line.
The ERBC approach shows strongly overconfident behavior in all cases since it is based on the error-rates of the whole system rather on single comparisons.
The DTC approach simply uses the distance between the score and the threshold as a confidence estimator.
Consequently, it overestimates less confident decisions and underestimates highly confident ones.
The LRC solution is based on the likelihood ratio between genuine and imposter.
Since this approach does not have a probabilistic interpretation, it strongly underestimates confidence.
The PFES and UMPC both require uncertainties per feature to make a confidence estimation.
Consequently, they could be only applied to PFE since this is the only utilized FRS able to state the uncertainty per feature.
UPMC strongly underestimates confidence for Adience and ColorFeret.
For LFW, the confidence estimates work well on average but are quite unstable.
This might be explainable through the training data.
Since the confidences are based on the uncertainties of the FRS.
The training data of the FRS might contain many well-illuminated and frontal faces, similar to LFW and unsimilar to Adience and ColorFeret.
For PFES, a similar behavior to DTC is observed, a mixture of strong over- and underestimating confidence.
Contrarily, the proposed PIC approach produces stable and accurate confidence estimations that are often close to the optimal solution (black line).

%table with ECE, MCE
\begin{table*}[ht]
\small
\renewcommand{\arraystretch}{1.0}
\setlength{\tabcolsep}{2.8pt}
\centering
\caption{\textbf{Confidence Calibration Analysis} - The ECE and MCE are shown for several dataset and FRS combinations at an FMR of $10^{-3}$.
ECE shows the average confidence calibration error, while the MCE presents the maximum calibration error. The best performance is marked in bold.
Except for one case, the proposed PIC approach strongly outperforms the other confidence estimators in terms of interpretability. This holds for the average performance, as well as for the worst-case scenario.}
\label{tab:CalibrationAnalysis}
\begin{tabular}{llrrrrrrrrrrrr}
\Xhline{2\arrayrulewidth}
           &          & \multicolumn{6}{c}{ECE [\%] at $10^{-3}$ FMR}  & \multicolumn{6}{c}{MCE [\%] at $10^{-3}$ FMR}    \\
           \cmidrule(rl){3-8} \cmidrule(rl){9-14} 
Database   & FRS      & PIC (ours) & ERBC  & DTC   & LRC   & UPMC & PFES  & PIC (ours) & ERBC  & DTC   & LRC   & UPMC   & PFES \\
\hline
Adience    & FaceNet  & \textbf{2.90}       & 26.19 & 23.45 & 24.68 & -    & -     & \textbf{15.55}      & 56.40 & 41.56 & 97.62 & -      & -     \\
           & PFE      & \textbf{1.14}       & 36.65 & 40.00 & 15.04 & 4.36 & 50.19 & \textbf{6.36}       & 85.07 & 48.87 & 93.06 & 90.83  & 57.73   \\
           & ArcFace  & \textbf{1.48}       & 38.53 & 20.89 & 19.31 & -    & -     & \textbf{18.00}      & 83.53 & 50.46 & 98.70 & -      & -     \\
           & MagFace  & \textbf{1.38}       & 40.13 & 22.43 & 18.73 & -    & -     & \textbf{13.00}      & 83.90 & 50.56 & 98.69 & -      & -     \\
           & QMagFace & \textbf{1.50}       & 39.25 & 31.79 & 18.73 & -    & -     & \textbf{13.29}      & 86.79 & 46.96 & 98.15 & -      & -     \\
\hline
ColorFeret & FaceNet  & \textbf{0.95}       & 40.83 & 26.68 & 14.12 & -    & -     & \textbf{8.33}       & 84.97 & 44.88 & 99.56 & -      & -     \\
           & PFE      & \textbf{0.96}       & 36.81 & 41.64 & 7.17  & 3.69 & 50.74 & \textbf{4.07}       & 75.30 & 45.78 & 92.33 & 80.91  & 54.83 \\
           & ArcFace  & \textbf{0.96}       & 46.44 & 16.99 & 4.68  & -    & -     & \textbf{10.91}      & 83.48 & 54.82 & 99.40 & -      & -     \\
           & MagFace  & \textbf{1.06}       & 46.44 & 18.91 & 4.79  & -    & -     & \textbf{8.01}       & 82.91 & 53.09 & 92.40 & -      & -     \\
           & QMagFace & \textbf{1.05}       & 45.39 & 34.00 & 4.02  & -    & -     & \textbf{6.32}       & 84.71 & 41.29 & 98.83 & -      & -     \\
\hline
LFW        & FaceNet  & \textbf{0.37}       & 40.76 & 30.43 & 11.01 & -    & -     & \textbf{15.76}      & 91.91 & 39.79 & 92.52 & -      & -     \\
           & PFE      & \textbf{0.07}       & 0.59  & 41.42 & 0.59  & 0.91 & 51.36 & \textbf{22.24}      & 94.50 & 48.87 & 93.07 & 49.39  & 56.37 \\
           & ArcFace  & \textbf{0.89}       & 40.58 & 19.15 & 12.04 & -    & -     & 54.99      & 91.22 & \textbf{47.24} & 92.26 & -      & -     \\
           & MagFace  & \textbf{0.09}       & 52.61 & 20.63 & 0.15  & -    & -     & \textbf{23.29}      & 92.87 & 49.76 & 98.89 & -      & -     \\
           & QMagFace & \textbf{0.07}       & 53.25 & 33.60 & 0.16  & -    & -     & \textbf{29.89}      & 92.35 & 52.85 & 99.45 & -      & -     \\
\hline
Morph      & FaceNet  & \textbf{0.23}       & 41.76 & 30.25 & 7.84  & -    & -     & \textbf{11.83}      & 90.90 & 40.44 & 92.56 & -      & -     \\
           & PFE      & \textbf{0.18}       & 50.88 & 44.02 & 0.94  & 1.34 & 49.50 & \textbf{32.34}      & 90.68 & 49.18 & 95.61 & 49.88  & 53.48 \\
           & ArcFace  & \textbf{0.06}       & 48.41 & 21.61 & 0.10  & -    & -     & \textbf{18.06}      & 91.19 & 48.24 & 99.14 & -      & -     \\
           & MagFace  & \textbf{0.23}       & 47.76 & 20.99 & 1.00  & -    & -     & \textbf{34.93}      & 92.36 & 48.71 & 99.19 & -      & -     \\
           & QMagFace & \textbf{0.83}       & 47.92 & 38.91 & 1.10  & -    & -     & \textbf{36.13}      & 91.40 & 55.83 & 98.59 & -      & -     \\
\Xhline{2\arrayrulewidth}
\end{tabular}
\end{table*}

To analyze the probabilistic interpretability quantitatively, Table \ref{tab:CalibrationAnalysis} shows the ECE and MCE of the confidence estimation methods for all database and FRS combinations at a FMR of $10^{-3}$.
The ECE shows how much a confidence estimator is off on average and thus, states how reliable a method can state confidence.
To also cover the worst-case scenarios, the MCE shows how much a confidence estimator is off in the worst-case.
For nearly all cases, the proposed PIC approach leads to significantly smaller ECE and MCE values demonstrating its effectiveness for estimating probabilistic interpretable confidence values.

\subsection{Multi-Comparison Analysis}
\label{sec:MultiComparisonAnalysis}
\vspace{-2mm}
\subsubsection{Recognition Performance}
\label{sec:MultiRecognition}
\vspace{-2mm}
The proposed PIC approach is able to naturally combine multiple samples into a single comparison score.
This is known as a multi-biometric fusion and aims to fuse information from multiple sources to improve recognition performance whilst addressing some of the limitations of single-biometric systems, such as poor data quality \cite{DBLP:conf/cvpr/TerhorstKDKK20} or overlap between identities \cite{DBLP:journals/inffus/SinghSR19}.
In this section, we will show that PIC significantly increases recognition performance with multiple samples.
In the next section, we will then demonstrate that the joint PIC scores also increase the probabilistic interpretability.

Table \ref{tab:MultiRecognitionAnalysis} shows the recognition performance of the proposed PIC approach when combining multiple samples.
The recognition error FNMR is shown for a fixed FMR of $10^{-3}$ for all database and FRS combinations.
In this multi-biometric context, a given probe sample is compared to 1/2/5 reference samples to calculate a joint PIC score.
The results demonstrate that the PIC score can be efficiently used for multi-biometric recognition scenarios.
{However, since the main contribution of this paper lies on probabilistic confidence estimation, it is not compared against (non-interpretable) score fusion approaches.}
In general, the recognition error for multiple reference samples is significantly lower than in the single-biometric case with one sample.
The only exceptions are mostly on the LFW dataset which is not well-suited for multi-biometric recognition analysis. 
For instance, 80\% of the identities in LFW have only one image.
Consequently, doing multi-sample comparisons results in an insignificant low number of genuine scores for the evaluation which makes the interpretation of the results less meaningful.
Besides this, the joint PIC score significantly increases the multi-biometric recognition performance in all other cases demonstrating its effectiveness for multi-biometrics besides confidence estimation.

\vspace{-3mm}
\subsubsection{Calibration Performance}
\vspace{-2mm}

In the following, we will show that the probabilistic confidence interpretation of the proposed PIC approach still remains for multi-biometric scenarios.
Table \ref{tab:MultiCalibrationAnalysis} analyses the probabilistic interpretability of the proposed PIC approach for the multi-biometric scenario.

For all database and FRS combinations, it shows the expected calibration errors (ECE) at a decision threshold for a $10^{-3}$ FMR.
Similar to before, a given probe sample is compared to 1/2/5 reference samples to calculate a joint PIC score confidence.
In general, confidence estimates of the proposed PIC approach improve when combining multiple samples.
In nearly all cases, the ECE decreases for more reference samples.
This demonstrates that, similar to the recognition performance, also the probabilistic confidence estimation of the joint PIC approach increases significantly.

\begin{table}[ht]
\small
\renewcommand{\arraystretch}{1.0}
\setlength{\tabcolsep}{7pt}
\centering
\caption{\textbf{Multi-Biometric PIC score recognition performance} - The recognition performance of using a joint PIC score by combining a probe sample with 1/2/5 reference samples (RS) is shown for all database and FRS combinations. Generally, the joint PIC score leads to lower recognition errors at $10^{-3}$ FMR than in the single-biometric scenario. 
}
\label{tab:MultiRecognitionAnalysis}
\begin{tabular}{llrrr}
\Xhline{2\arrayrulewidth}
           &          & \multicolumn{3}{c}{FNMR [\%] at $10^{-3}$ FMR} \\
           \cmidrule(rl){3-5} 
Database   & FRS      & \multicolumn{1}{c}{1 RS} &\multicolumn{1}{c}{2 RS} & \multicolumn{1}{c}{5 RS}   \\
\hline
Adience    & FaceNet  &71.00&65.81 & 58.77     \\
           & PFE      &18.47& 30.22 & 19.43   \\
           & ArcFace  &10.10&6.43 &2.57           \\
           & MagFace  &9.75&5.00  & 2.59      \\
           & QMagFace &9.78&5.09   & 2.28       \\
\hline
ColorFeret & FaceNet  & 12.22 &6.09 & 4.47         \\
           & PFE      &6.60& 8.66 & 9.08 \\
           & ArcFace  & 4.22 &2.18   & 2.39        \\
           & MagFace  &3.92&1.86   & 1.77          \\
           & QMagFace &3.24&1.47  & 1.42          \\
\hline
LFW\tablefootnote{The identity distribution of LFW does not allow creating many multi-sample comparisons. Thus, is not well-suited for analyzing multi-biometric recognition. However, we reported these results for the sake of completeness.}        & FaceNet  & 0.79 &1.02 & 2.05           \\
           & PFE      &0.08& 0.25 & 0.24\\
           & ArcFace  &4.38&2.31  & 1.88           \\
           & MagFace  &0.05&0.16   & 0.28            \\
           & QMagFace &0.05&0.18 & 0.27    \\
\hline
Morph      & FaceNet  & 0.54 &0.15 & 0.07         \\
           & PFE      & 0.89& 0.70 & 0.52\\
           & ArcFace  &0.05&0.05 & 0.03          \\
           & MagFace  &0.96&0.48  & 0.45   \\
           & QMagFace &0.96&0.50   & 0.42        \\
\Xhline{2\arrayrulewidth}
\end{tabular}
\end{table}

\begin{table}[ht]
\small
\renewcommand{\arraystretch}{1.0}
\setlength{\tabcolsep}{7pt}
\centering
%\footnotesize
\caption{\textbf{Multi-Biometric Confidence Calibration Analysis} - The expected calibration errors (ECE) are shown for a joint PIC score confidence by combining a probe sample with 1/2/5 reference samples (RS).
This was done for all database and FRS combinations at an FMR of $10^{-3}$.
The ECE significantly decreases when more reference samples are used. Consequently, also the probabilistic confidence interpretation of PIC becomes more accurate when multiple samples are combined.
}\label{tab:MultiCalibrationAnalysis}
\begin{tabular}{llrrr}
\Xhline{2\arrayrulewidth}
           &          & \multicolumn{3}{c}{ECE [\%]}     \\
           \cmidrule(rl){3-5}
Database   & FRS      & \multicolumn{1}{c}{1 RS}& \multicolumn{1}{c}{2 RS} & \multicolumn{1}{c}{5 RS}   \\
\hline
Adience    & FaceNet  &2.90&1.80 &0.44  \\
           & PFE      &1.14 & 0.51 &  0.00    \\
           & ArcFace  &1.48&0.26 &0.11\\
           & MagFace  &1.38&0.27&0.15\\
           & QMagFace &1.50&0.31&0.10\\
\hline
ColorFeret & FaceNet  &0.95& 0.40 &0.36\\
           & PFE     & 0.96 & 0.63 & 3.03    \\
           & ArcFace  &0.96& 0.24&0.06\\
           & MagFace &1.06&0.19&0.13\\
           & QMagFace &1.05&0.17&0.17\\
\hline
LFW        & FaceNet  &0.37& 0.09&0.11\\
           & PFE      & 0.07& 0.27 & 0.00 \\
           & ArcFace &0.89&0.07&0.01\\
           & MagFace &0.09&0.03&0.04\\
           & QMagFace &0.07&0.03&0.14    \\
\hline
Morph      & FaceNet  & 0.23&0.07  &0.02\\
           & PFE      &  0.18 & 0.16 & 0.00 \\
           & ArcFace &0.06&0.01  &0.00\\
           & MagFace  &0.23& 0.01 &0.00\\
           & QMagFace &0.83&0.03&0.00\\
\Xhline{2\arrayrulewidth}
\end{tabular}
\end{table}

%%%%%%%%% Conclusion
\section{Conclusion}
\label{sec:Conclusion}

Since mistakes are coming at high costs in critical decision-making processes, such as forensics or law enforcement, it is necessary to accurately state the matching confidence of a biometric system.
Previous works on confidence estimation in biometrics can well differentiate between low and high-confident decisions but produce confidence estimates that are not interpretable as the probability that the decision made is correct.
To fix this issue, we proposed the PIC score, a probabilistic interpretable comparison score.
The conducted experiments demonstrate that the proposed approach outperforms all related approaches in terms of probabilistic interpretability and can also be applied in multi-biometric recognition scenarios.
The proposed PIC score jointly achieves interpretability, optimality, universality, combinability, and integratability.
Moreover, the score accurately states the probability that the compared samples belong to the same identity (interpretability).
Since it was derived from the Bayes' theorem, it provides optimal matching confidences given suitable training data (optimality).
It can be applied to any biometric modality and system without changing its single-biometric performance (universality).
In contrast to the standard comparison score, multiple samples can be efficiently combined into a joint PIC score (combinability) leading to a significant gain in recognition performance and interpretability.
Lastly, the PIC solution can be easily integrated into existing biometric systems and avoids the need for data- and computationally-expensive experiments to determine the desired decision threshold.

\footnotesize{
\section*{Acknowledgement}

This work is co-financed by Component 5 - Capitalization and Business Innovation, integrated in the Resilience Dimension of the Recovery and Resilience Plan within the scope of the Recovery and Resilience Mechanism (MRR) of the European Union (EU), framed in the Next Generation EU, for the period 2021 - 2026, within project NewSpacePortugal, with reference 11, and by National Funds through the Portuguese funding agency, FCT - Fundação para a Ciência e a Tecnologia within the PhD grant ‘‘2021.06872.BD’’.
Portions of the research in this paper use the FERET database of facial images collected under the FERET program, sponsored by the DOD Counterdrug Technology Development Program Office. }

%\clearpage
%\newpage

%%%%%%%%% REFERENCES
{\small
\bibliographystyle{ieee_fullname}
\bibliography{egbib}
}

\end{document}